\documentclass[runningheads]{llncs}
\usepackage{graphicx}
\usepackage{tikz}
\usepackage[caption=false]{subfig}
\usepackage{amssymb}
\usepackage{amsmath}
\usepackage{subfig}
\usetikzlibrary{
    shapes.geometric,
    decorations.pathreplacing,
    angles,quotes,intersections
}
\tikzset{
    elli/.style args={#1:#2and#3}{
        draw,
        shape=ellipse,
        rotate=#1,
        minimum width=2*#2,
        minimum height=2*#3,
        outer sep=0pt,
    },
    /pgf/decoration/raise/.append code={
        \def\tikzdecorationsbrace{#1}
    },
    elli node/.style={
        circle,
        black,
        draw=none,
        midway,
        anchor=#1-90,
        inner sep=0pt,
        shift=(#1+90:\tikzdecorationsbrace+\pgfdecorationsegmentamplitude)
    },
    eigen/.style 2 args={
        decorate,
        decoration={
            brace,
            amplitude=#1,
            mirror,
            raise=#2,
        },
    },
    eigen/.default={15pt}{4pt},
    axis/.style={
        line width=.5mm,
        ->,
    },
    normal axis/.style={
        axis,
        dashed,
    }
}

\usepackage{xspace}

\usepackage[normalem]{ulem}

\begin{document}
\title{A Bayesian Convolutional Neural Network for Robust Galaxy Ellipticity Regression}
\titlerunning{A Bayesian CNN for Robust Galaxy Ellipticity Regression}
%

\author{Claire Theobald\inst{1} \and
Bastien Arcelin\inst{2} \and
Frédéric Pennerath\inst{1} \and
Brieuc Conan-Guez\inst{1}  \and
Miguel Couceiro\inst{1} \and
Amedeo Napoli\inst{1}
}

\institute{
Université de Lorraine, CentraleSupélec, CNRS, Inria, LORIA, F-54000 France \and
Université de Paris, CNRS, Astroparticule et Cosmologie, F-75013 Paris, France
}

\authorrunning{C. Theobald et al.}

%
%
\maketitle              
\begin{abstract}
        Cosmic shear estimation is an essential scientific goal for large galaxy surveys. It refers to the coherent distortion of distant galaxy images due to weak gravitational lensing along the line of sight. It can be used as a tracer of the matter distribution in the Universe. The unbiased estimation of the local value of the cosmic shear can be obtained via Bayesian analysis which relies on robust estimation of the galaxies ellipticity (shape) posterior distribution. This is not a simple problem as, among other things, the images may be corrupted with strong background noise.  For current and coming surveys, another central issue in galaxy shape determination is the treatment of statistically dominant overlapping (blended) objects. We propose a Bayesian Convolutional Neural Network based on Monte-Carlo Dropout to reliably estimate the ellipticity of galaxies and the corresponding measurement uncertainties. We show that while a convolutional network can be trained to correctly estimate well calibrated aleatoric uncertainty, -the uncertainty due to the presence of noise in the images- it is unable to generate a trustworthy ellipticity distribution when exposed to previously unseen data (i.e. here, blended scenes). By introducing a Bayesian Neural Network, we show how to reliably estimate the posterior predictive distribution of ellipticities along with robust estimation of epistemic uncertainties. Experiments also show that epistemic uncertainty can detect inconsistent predictions due to unknown blended scenes.
        
\end{abstract}

\section{Introduction}
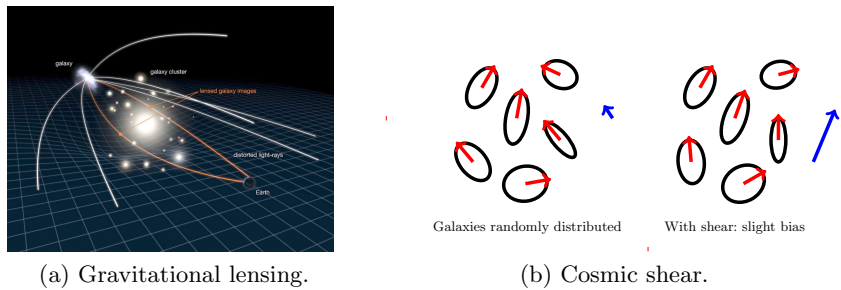
\begin{figure}[htbp]
\begin{center}
\subfloat[Gravitational lensing.\label{fig:grav_lens}]{%
\includegraphics[scale=0.12]{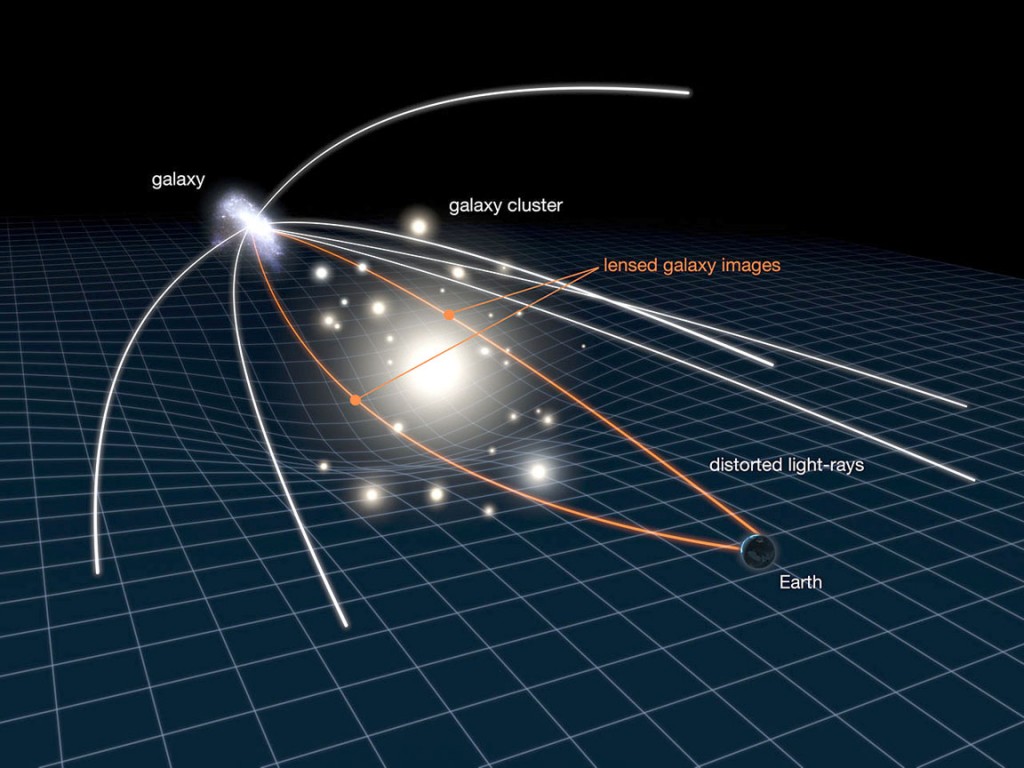} 
}\hfil
\subfloat[Cosmic shear.\label{fig:weak_lens}]{%
\scalebox{.58}{
\begin{tikzpicture}
\node[elli=80:0.6cm and 0.25cm, line width = 0.8mm] at (-3, 0) (e) {};
\draw [->, color = red, line width = 0.8mm] (-3, 0) -- (-2.879, 0.689);
\node[elli=10:0.5cm and 0.40cm, line width = 0.8mm] at (-2.8, -1.5) (e) {};
\draw [->, color = red, line width = 0.8mm] (-2.8, -1.5) -- (-2.2092, -1.3959);
\node[elli=130:0.5cm and 0.17cm, line width = 0.8mm] at (-2, -0.5) (e) {};
\draw [->, color = red, line width = 0.8mm] (-2, -0.5) -- (-2.386, -0.04);
\node[elli=60:0.5cm and 0.28cm, line width = 0.8mm] at (-3.8, 0.7) (e) {};
\draw [->, color = red, line width = 0.8mm] (-3.8, 0.7) -- (-3.5, 1.219);
\node[elli=130:0.5cm and 0.3cm, line width = 0.8mm] at (-4, -1) (e) {};
\draw [->, color = red, line width = 0.8mm] (-4, -1) -- (-4.386, -0.5404);
\node[elli=155:0.4cm and 0.3cm, line width = 0.8mm] at (-2, 1) (e) {};
\draw [->, color = red, line width = 0.8mm] (-2, 1) -- (-2.453, 1.211);

\draw [->, color = blue, line width = 0.8mm] (-0.8, 0) -- (-1, 0.3);

\node[] at (-3,-2.5) {\hspace{0.5cm}Galaxies randomly distributed};

\node[elli=70:0.6cm and 0.25cm, line width = 0.8mm] at (2, 0) (e) {};
\draw [->, color = red, line width = 0.8mm] (2, 0) -- (2.239, 0.6578);
\node[elli=30:0.5cm and 0.40cm, line width = 0.8mm] at (2.2, -1.5) (e) {};
\draw [->, color = red, line width = 0.8mm] (2.2, -1.5) -- (2.720, -1.2);
\node[elli=90:0.5cm and 0.17cm, line width = 0.8mm] at (3, -0.5) (e) {};
\draw [->, color = red, line width = 0.8mm] (3, -0.5) -- (3, 0.1);
\node[elli=60:0.5cm and 0.28cm, line width = 0.8mm] at (1.2, 0.7) (e) {};
\draw [->, color = red, line width = 0.8mm] (1.2, 0.7) -- (1.5, 1.22);
\node[elli=95:0.5cm and 0.3cm, line width = 0.8mm] at (1, -1) (e) {};
\draw [->, color = red, line width = 0.8mm] (1, -1) -- (0.948, -0.403);
\node[elli=15:0.4cm and 0.3cm, line width = 0.8mm] at (3, 1) (e) {};
\draw [->, color = red, line width = 0.8mm] (3, 1) -- (3.483, 1.129);

\draw [->, color = blue, line width = 0.8mm] (3.8, -1) -- (4.3, 0.2);
\node[] at (2,-2.5) {With shear: slight bias};

\draw[line width = 1mm, color=red] (0,-3) -- (0,-3); {};
\draw[line width = 1mm, color=red] (-6,0) -- (-6,0); {};
\end{tikzpicture}
}
}\hfil
\caption{\protect\subref{fig:grav_lens} Effect of gravitational lensing: the mass bends the light and deforms the images of the galaxies. \protect\subref{fig:weak_lens} Weak lensing: the correlation between orientations and shapes of neighbour galaxies defines the cosmic shear. In blue: average ellipticity. Left: the expected ellipticity distribution. Right: the observed ellipticity distribution. Image: \protect\subref{fig:grav_lens} NASA/ESA}
\label{fig:grav_weak_lens}
\end{center}
\end{figure}

One of the goals of large galaxy surveys such as the \emph{Legacy Survey of Space and Time} (LSST, \cite{2009arXiv0912.0201L}) conducted at the Vera C. Rubin Observatory is to study \emph{dark energy}. This component of unknown nature was introduced in the current cosmological standard model to explain the acceleration of the Universe expansion. One way to probe dark energy is to study the mass distribution across the Universe. This distribution mostly follows the dark matter distribution, which does not interact with baryonic matter (i.e. visible matter) except through gravitation, as dark matter represents around 85\% of the matter in the Universe.
Consequently, cosmologists need to use indirect measurement techniques such as \emph{cosmic shear}, which measures the coherent distortion of background galaxies images by foreground matter due to \emph{weak gravitational lensing} \cite{kilbingercosmicshear}. In astrophysics, gravitational lensing is the distortion of the image of an observed source, induced by the bending of space-time, thus of the light path, generated by the presence of mass along the line of sight.
The mass acts like a lens, in partial analogy with optical lenses, as illustrated in Fig.\ref{fig:grav_lens}. The weak gravitational lensing effect is faint (1\% of galaxy shape measurement) and only statistical tools provide a way to detect a local correlation in the observed galaxies orientations. This correlation yields a local value at every point of the observable Universe, defining the cosmic shear field. As pictured in Fig.\ref{fig:weak_lens}, in an isotropic and uniform Universe orientations of galaxies are expected to follow a uniform distribution (left panel). The statistical average of their oriented elongations, hereafter called \emph{complex ellipticities}, is expected to be null. In presence of a lens, a smooth spatial deformation field modifies coherently the complex ellipticities of neighboring galaxies so that their mean is no longer zero (right panel).

The unbiased measurement of cosmic shear is a major ambition of nowadays cosmology \cite{mandelbaumwl}. One avenue to estimate the cosmic shear locally is to combine individual galaxy ellipticity measurements. By looking deeper into the sky, that is to older objects, the next generation of telescopes will allow for the detection of a very large number of galaxies, potentially leading to very precise shear measurement and resulting in tight constraints on dark energy parameters. 

Methods already exist to estimate galaxy ellipticities through direct measurement on images recorded by telescope cameras (\cite{ref_kaiser_ksb_1995} for example). This is a complex problem as, among other things, the shear signal is carried by faint galaxies which makes it very sensitive to background noise. Another central issue for current and coming surveys in galaxy shape determination, is the treatment of statistically dominant overlapping objects, an effect called \emph{blending}. A current survey projects that 58\% of the detected objects will appear blended \cite{2018PASJ...70S...5B} and this value is expected to reach around 62\% for LSST \cite{2021arXiv210302078S}. To overcome this issue, solutions exist such as deblending \cite{sextractor,scarlet,deblendervae}: the separation of overlapping objects. Yet, they are not perfect and rely on an accurate detection of blended scenes which is also a complex problem. As such, in addition to a precise estimation of the complex ellipticities, a reliable measurement of the uncertainties is crucial in order to discard, or at least decrease the impact of, unreliable and inaccurate measurements avoiding as much as possible the introduction of a bias into the shear estimation.

Classical ellipticity measurement methods usually adopt assumptions about the shape of the galaxies (for example via the shape of the window function in \cite{ref_kaiser_ksb_1995}) potentially resulting in model bias. In contrast, \emph{convolutional neural networks} or CNNs \cite{ref_lecun_dl_2015} make it possible to learn and recognize complex and diverse galaxy shapes directly from data without making any other hypothesis than the representativeness of the training sample. They consequently are appropriate tools to learn the regression of galaxy ellipticities, even in the presence of noise and complex distortions.
Yet standard CNNs can only measure the \emph{aleatoric uncertainty}: the one due to the presence of noise in the data. They are unable to estimate the \emph{epistemic uncertainty}, the one due to the limited number of samples a CNN has been trained with and to the model \cite{ref_gal_dropout_2016,DBLP:conf/icml/LeSC05}. This second type of uncertainty is essential to detect outliers from the training samples, or formulated accordingly to our problem, to distinguish between reliable or unreliable galaxy ellipticity estimation. It is only accessible by considering neural network weights as random variables instead of constants, that is, by adopting a Bayesian approach. Consequently, we have focused our work on \emph{Bayesian Deep Learning} \cite{ref_hinton_keeping_1993} using Monte Carlo dropout (MC dropout) \cite{ref_gal_dropout_2016} as the mean to apply Bayesian inference to Deep Learning models.  

Foreseeing a Bayesian estimation of the cosmic shear, combining galaxy ellipticity posteriors estimated directly from images (with blends or not) in different filters (or bands), this paper focuses on estimating reliable galaxy ellipticity posteriors from single band images. 
This is a necessary step to check that the proposed method efficiently estimates a calibrated aleatoric uncertainty and is able to minimize the impact of wrongly estimated ellipticity values due to outliers in the computation of the shear. We compare two networks trained on isolated galaxy images with or without noise in order to test for the calibration of aleatoric uncertainty. Regarding outliers, blended scenes are perfect examples. Note that these are illustrations of aleatoric or epistemic uncertainty sources. Most of cosmic shear bias sources such as detection, Point-Spread-Function (PSF) treatment, or selection for example \cite{kilbingercosmicshear,mandelbaumwl}, can fall in one or the other category. The estimation of galaxy ellipticity posterior from blended scenes in different bands is a harder problem that we will investigate in further work.


The contributions of this article are 1) to propose a Bayesian Deep Learning model that solves a complex multivariate regression problem of estimating the galaxy shape parameters while accurately estimating aleatoric and epistemic uncertainties; 2) to establish an operational protocol to train such a model based on multiple incremental learning steps; and 3) to provide experimental evidences that the proposed method is able to assess 
whether an ellipticity measurement is reliable. This is illustrated, in this paper, by the accurate differentiation between isolated galaxy or blended scenes, considered here as outliers, and the relationship between epistemic uncertainty and predictive ellipticity error. We also show that this last result could not be obtained with a classical, non Bayesian network.

The rest of the paper is organized as follows. In Section~\ref{sec:prob} we briefly describe the problem to be solved and comment on some of its peculiarities. We detail our proposed  solution in Section~\ref{sec:sol}. We analyse the results obtained on the various experiments we performed in  Section~\ref{sec:res}, and we conclude and give the directions of further research in Section~\ref{sec:conclusion}.

\section{Estimating galaxy ellipticity from images}
\label{sec:prob}
As mentioned previously, it is possible to estimate cosmic shear combining individual measurements of galaxy shape. 
This shape information can be quantified by the complex ellipticity, which can be defined in cosmology as in Def.~\ref{def:complex_ellipticity}.
\begin{definition}\label{def:complex_ellipticity}
Let $E$ be an ellipse with major axis $a$, minor axis $b$,  and with $\theta$ as its position angle.  The \emph{complex ellipticity} of $E$ is defined as:
\begin{equation}
    \epsilon = \epsilon_1 + \epsilon_2 \,i= \frac{1-q^2}{1+q^2}\, e^{2i\theta},
    \label{eq:complex_ellipticity}
\end{equation}
where  $q=\frac{b}{a}$ is the axis ratio of the ellipse.
\end{definition}

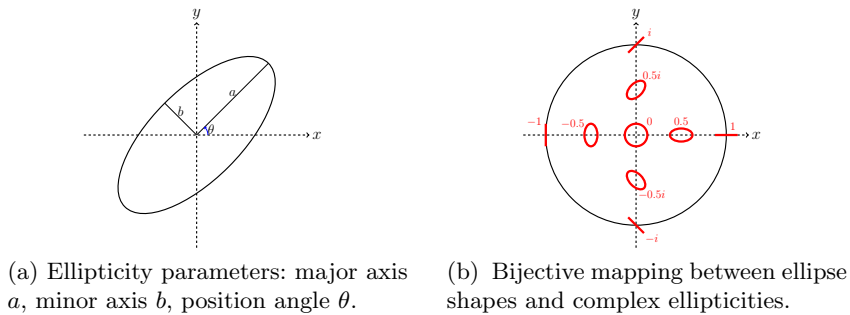
\begin{figure}[h]
\begin{center}
\subfloat[Ellipticity parameters: major axis $a$, minor axis $b$, position angle $\theta$\label{fig:ellipse_parameters}.]{%
\quad\quad\quad
\scalebox{0.30}{
\begin{tikzpicture}
\node[elli=45:4.5cm and 2.0cm, line width = 0.5mm] at (0, 0) (e) {};
\draw[normal axis, name path=A] (-5, 0) coordinate (A) -- (5, 0) coordinate (B) node[right] {\huge $x$};
\draw[normal axis, name path=B] (0, -5) -- (0, 5) node[above] {\huge $y$};

\draw[thick] ([shift={(45:0cm)}]    e.center) coordinate (C) -- ([shift={(45:4.5cm)}]    e.center)  coordinate (D) node[midway, above] {\LARGE $a$};
\draw[thick] ([shift={(90+45:0cm)}] e.center) -- ([shift={(90+45:2cm)}] e.center) node[midway, above]  {\LARGE $b$};
\path[name intersections={of=A and B,name=inter}];
\draw pic["\LARGE $\theta$", ->, draw=blue, angle eccentricity=1.4] {angle=B--inter-1--D};
\end{tikzpicture}
}
\quad\quad\quad
}\hfil
\subfloat[
Bijective mapping between ellipse shapes and complex ellipticities.\label{fig:complex_bijection}]{%
\quad\quad\quad
\scalebox{0.30}{
\begin{tikzpicture}
\draw[normal axis, name path=A] (-5, 0) coordinate (A) -- (5, 0) coordinate (B) node[right] {\huge $x$};
\draw[normal axis, name path=B] (0, -5) -- (0, 5) node[above] {\huge $y$};

\node[elli=45:4cm and 4cm, line width = 0.5mm] at (0, 0) (e) {};
\draw[line width = 1mm, color=red] (0.6,0.6) node{\Large $0$};
\node[elli=45:0.5cm and 0.5cm, line width = 1mm, color=red] at (0, 0) (e) {};
\draw[line width = 1mm, color=red] (2,0.6) node{\Large $0.5$};
\node[elli=0:0.5cm and 0.289cm, line width = 1mm, color=red] at (2, 0) (e) {};
\draw[line width = 1mm, color=red] (4.3,0.4) node{\Large $1$};
\draw[line width = 1mm, color=red] (3.5,0) -- (4.5,0); {};

\draw[line width = 1mm, color=red] (-2.8,0.5) node{\Large $-0.5$};
\node[elli=90:0.5cm and 0.289cm, line width = 1mm, color=red] at (-2, 0) (e) {};
\draw[line width = 1mm, color=red] (-4.5,0.6) node{\Large $-1$};
\draw[line width = 1mm, color=red] (-4,-0.5) -- (-4,0.5); {};

\draw[line width = 1mm, color=red] (0.7,2.65) node{\Large $0.5i$};
\node[elli=45:0.5cm and 0.289cm, line width = 1mm, color=red] at (0, 2) (e) {};
\draw[line width = 1mm, color=red] (0.6,4.55) node{\Large $i$};
\draw[line width = 1mm, color=red] (-0.35,3.65) -- (0.35,4.35); {};

\draw[line width = 1mm, color=red] (0.7,-2.65) node{\Large $-0.5i$};
\node[elli=-45:0.5cm and 0.289cm, line width = 1mm, color=red] at (0, -2) (e) {};
\draw[line width = 1mm, color=red] (0.7,-4.6) node{\Large $-i$};
\draw[line width = 1mm, color=red] (-0.35,-3.65) -- (0.35,-4.35); {};
\end{tikzpicture}
}
\quad\quad\quad
}\hfil
\end{center}
\caption{Geometric representation of the complex ellipticity. \protect\subref{fig:ellipse_parameters} The ellipse parameters. \protect\subref{fig:complex_bijection} The complex ellipticity defines a bijection between ellipse shapes and the unit disk. An ellipticity with low magnitude is close to a circle, while one with a high magnitude is closer to a straight line. The argument defines the orientation of the ellipse.}
\label{fig:complex_ellipticity}
\end{figure}

An illustration of the ellipticity parameters is shown in Fig.~\ref{fig:ellipse_parameters}. The complex ellipticity defines a bijection between the orientation and the elongation of the ellipse on one side, and the unit disk on the other side, see Fig.~\ref{fig:complex_bijection}.

\begin{figure}[h]
\centering
\subfloat[Isolated noiseless galaxy\label{fig:galim1}]{%
\quad\quad
\includegraphics[scale=0.27, trim=80 13 80 40, clip]{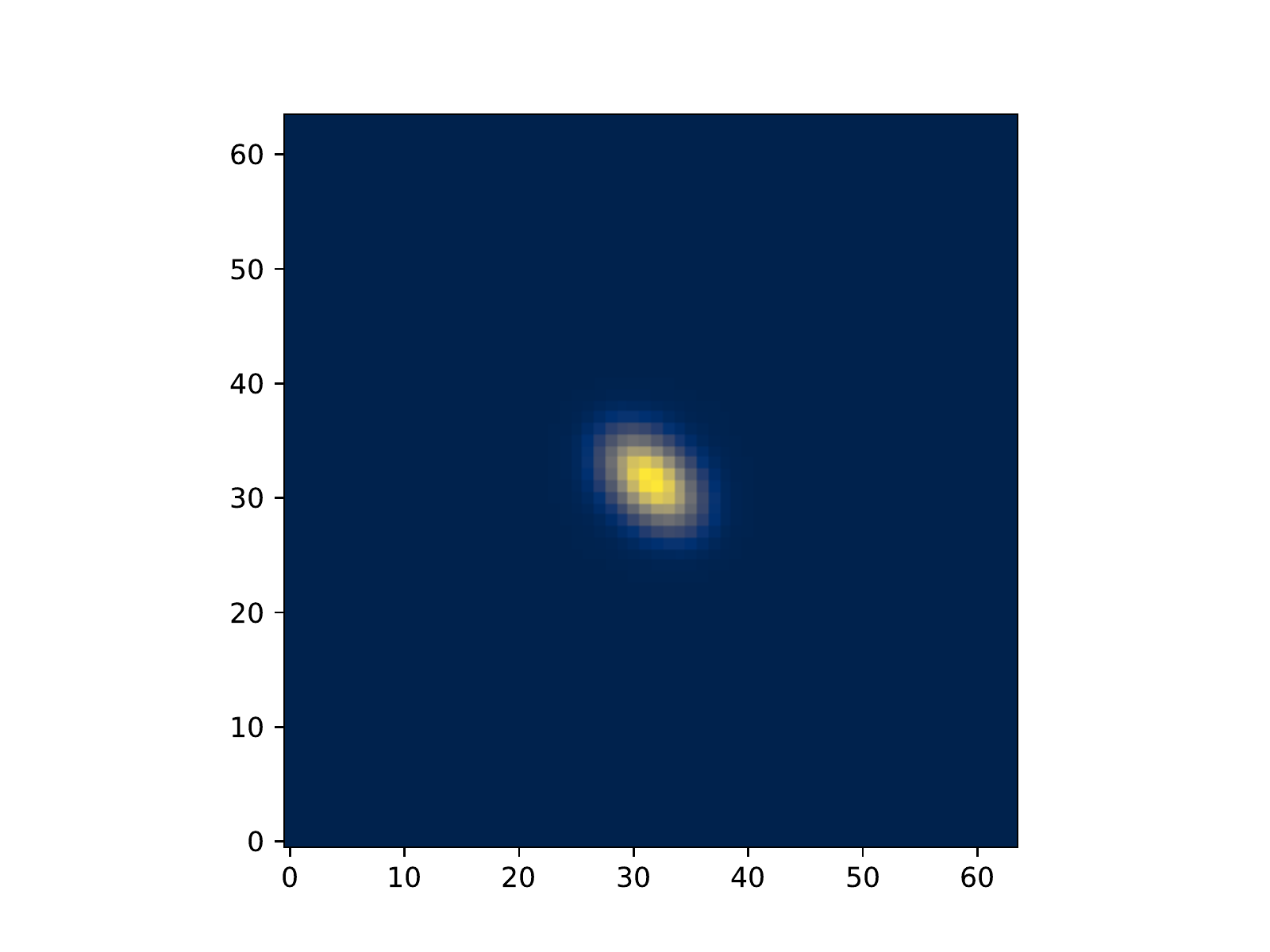} 
\quad\quad
}\hfil
\subfloat[Isolated noisy galaxy\label{fig:galim2}]{%
\quad
\includegraphics[scale=0.27, trim=80 13 80 40, clip]{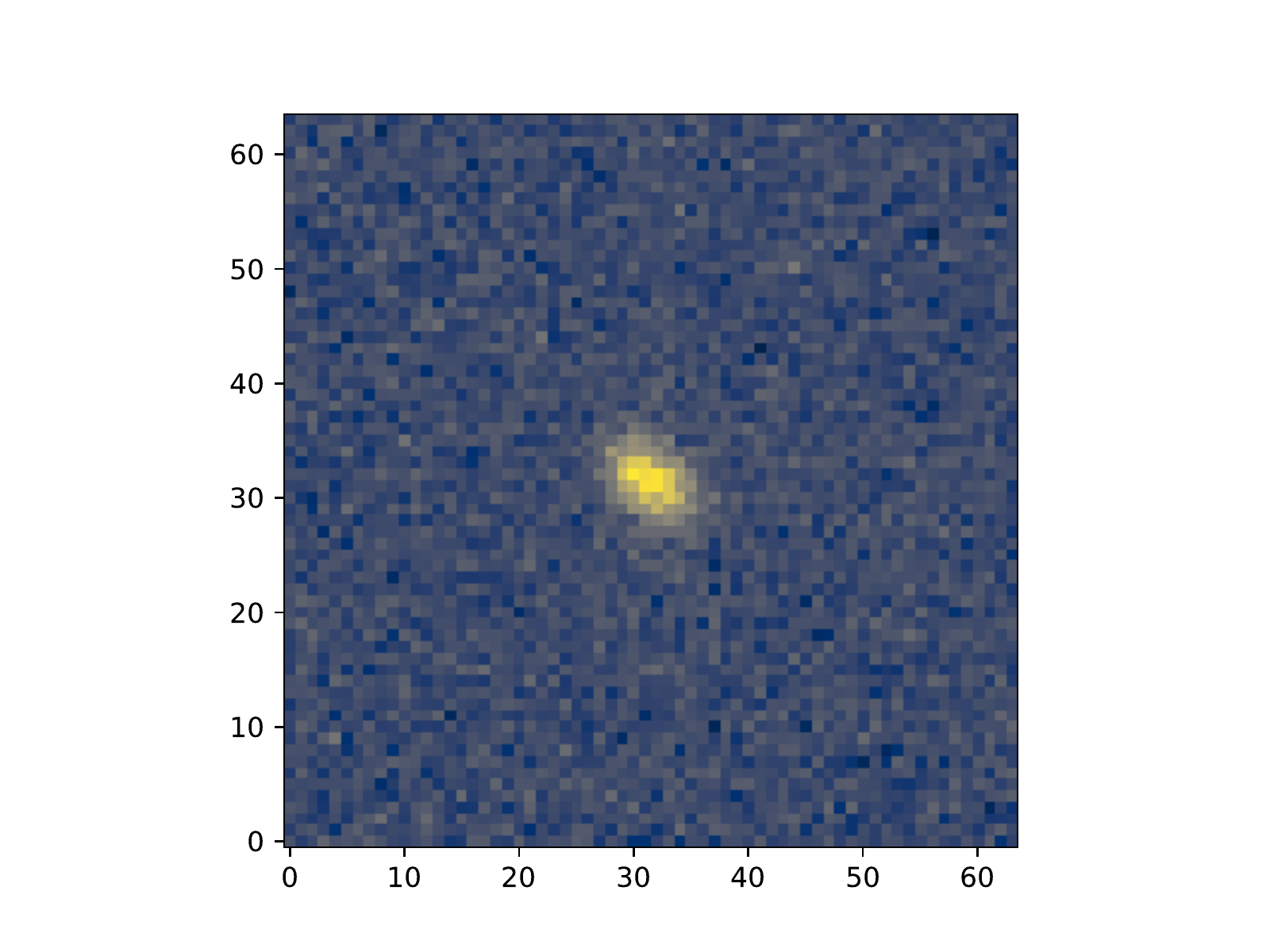}
\quad
}\hfil
\subfloat[Blended noisy galaxies\label{fig:galim3}]{%
\quad
\includegraphics[scale=0.27, trim=80 13 80 40, clip]{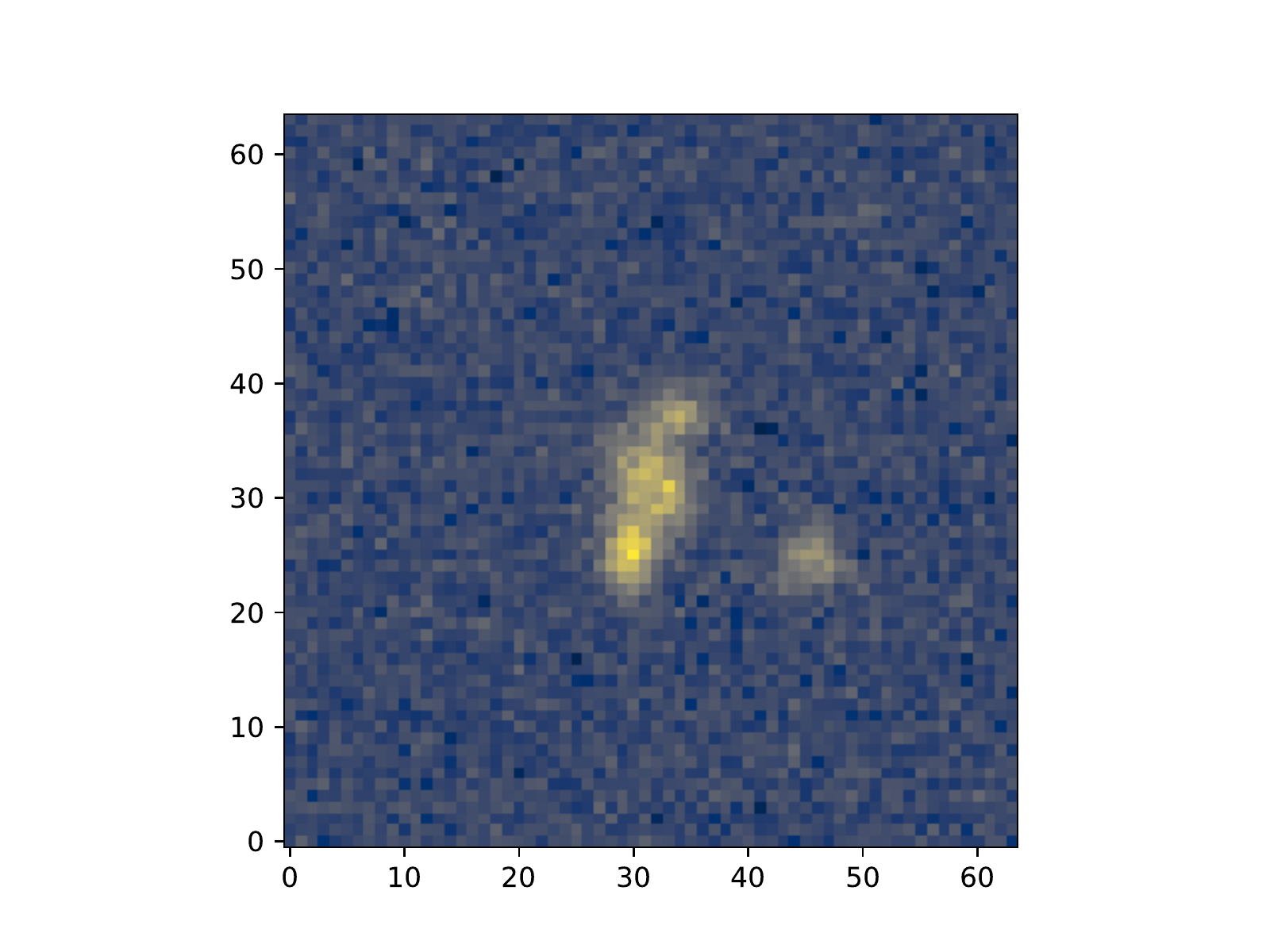} 
\quad
}\hfil

\caption{Three different types of image complexity for the same galaxy: isolated without noise, isolated with noise, blended with noise. Notice how the noise slightly deforms the galaxy \protect\subref{fig:galim2} and how the blended galaxies makes the ellipticity estimation very difficult \protect\subref{fig:galim3} when compared to a simple isolated galaxy without noise \protect\subref{fig:galim1}.}
\label{fig:galaxy_types}
\end{figure}

However, the process to achieve an unbiased measurement of cosmic shear, starting with the estimation of ellipticities, is going to be challenging for several reasons. We test the reliability of our networks prediction on noise and blending, two of the many possible bias sources in the cosmic shear estimation. Both of these issues result from the fact that the shear signal is mostly carried by faint galaxies.
By definition, these objects have a low signal-to-noise ratio. The noise corrupts the galaxy images, making the shape estimation much harder (see Fig. \ref{fig:galim2}), and can introduce a bias in shear measurement \cite{kilbingercosmicshear}. 

Also, a large part of these faint objects will appear blended with foreground galaxies. Even in scenes where objects are only slightly overlapped, the apparent shape of the detected object does not correspond to a single galaxy model and an ellipticity measurement on this image could give a completely wrong result.
Again, this work is the first step of a longer-term goal. Here, we target a reliable estimation of galaxy ellipticity posterior from single band images. This includes obtaining a well calibrated aleatoric uncertainty, tested here with and without the addition of Poisson noise on images, and an epistemic uncertainty allowing for minimization of the impact of untrustworthy measurement due to outliers (here, blended scenes). 
We simulate LSST-like images, allowing us to control the parameters of the scene, e.g., the number of galaxies, their location on the image, and the level and type of noise applied. We consider four categories of simulated data: isolated centered galaxies without noise, isolated centered galaxies with noise, 
and blended scene with and without noise. Images are 64x64 pixels stamps simulated in the brightest of the six bands corresponding to the LSST filters, each of them selecting a different part of the electromagnetic spectrum. These images are simulated placing, in their center, a galaxy whose ellipticity is to be measured.

The image generating process relies on the GalSim library \cite{rowe2015galsim} and is based on a catalog of parametric models fitted to real galaxies for the third Gravitational Lensing Accuracy Testing (GREAT3) Challenge \cite{2014ApJS..212....5M}. It consists in 1) producing an image of a centered noiseless isolated galaxy from a model sampled randomly from the catalog, with its corresponding physical properties (size, shape, orientation, PSF, brightness, redshift, etc) 2) measuring the complex ellipticity of the galaxy with the KSB algorithm \cite{ref_kaiser_ksb_1995} on the image 
and record it as the image label, 3) possibly adding on random image location other galaxy images (from 0 to 5) to generate blended scenes 4) possibly adding Poisson noise (as in \cite{deblendervae}).
In this study and for sake of interpretability, we only provide as input to our CNN the reference band (the brightest) which we use to define the target ellipticity, making our images two-dimensional. Once again, while using multiple bands is useful for blended galaxies \cite{scarlet,deblendervae}, here we focus only on predicting the ellipticity of a single centered galaxy with a correctly estimated uncertainty.

\section{A method to assess uncertainty in ellipticity estimation \label{sec:sol}}

\subsection{Estimation of noise related uncertainty} \label{sec:dlr}

Our first goal is to reliably estimate the first layer of complexity in the galaxy images, the noise. Given the nature of the data, we will be using a CNN \cite{ref_lecun_dl_2015}. However, training a CNN to solve a standard regression problem with an L2 loss does not allow us to estimate the uncertainty due to the noise. Therefore, in place of a complex scalar output, we predict a 2D \emph{multivariate normal distribution} (MVN) as an output: given an input image $X$, whose complex ellipticity is denoted $Y$ and given weight parameters $w$, the network outputs an MVN $Y \sim \mathcal{N}(\mu(X,w),\Sigma(X,w))$. As such, the model is no longer trained on a simple L2 loss but rather on the log-likelihood of the MVN. The mean of the distribution $\mu(X,w)$, which is also the mode, serves as the predicted output, and the covariance matrix $\Sigma(X,w)$, which is also an output of the network, represents the so-called aleatoric uncertainty on the input data $X$. The model is therefore heteroscedastic, as $\Sigma(X,w)$ depends on the input $X$ \cite{DBLP:conf/icml/LeSC05}. This allows our model to estimate the aleatoric uncertainty for each image individually. The determinant of $\Sigma(X,w)$, denoted $|\Sigma(X,w)|$, is a scalar measure of uncertainty, as it is directly related to the differential entropy, $\ln\left(\sqrt{(2\pi e)^2|\Sigma(X,w)|}\right)$, of an MVN. 

\begin{figure}[htp]
\centering
\subfloat[Convolutional layers\label{fig:nnim1}]{%
\includegraphics[scale=0.27]{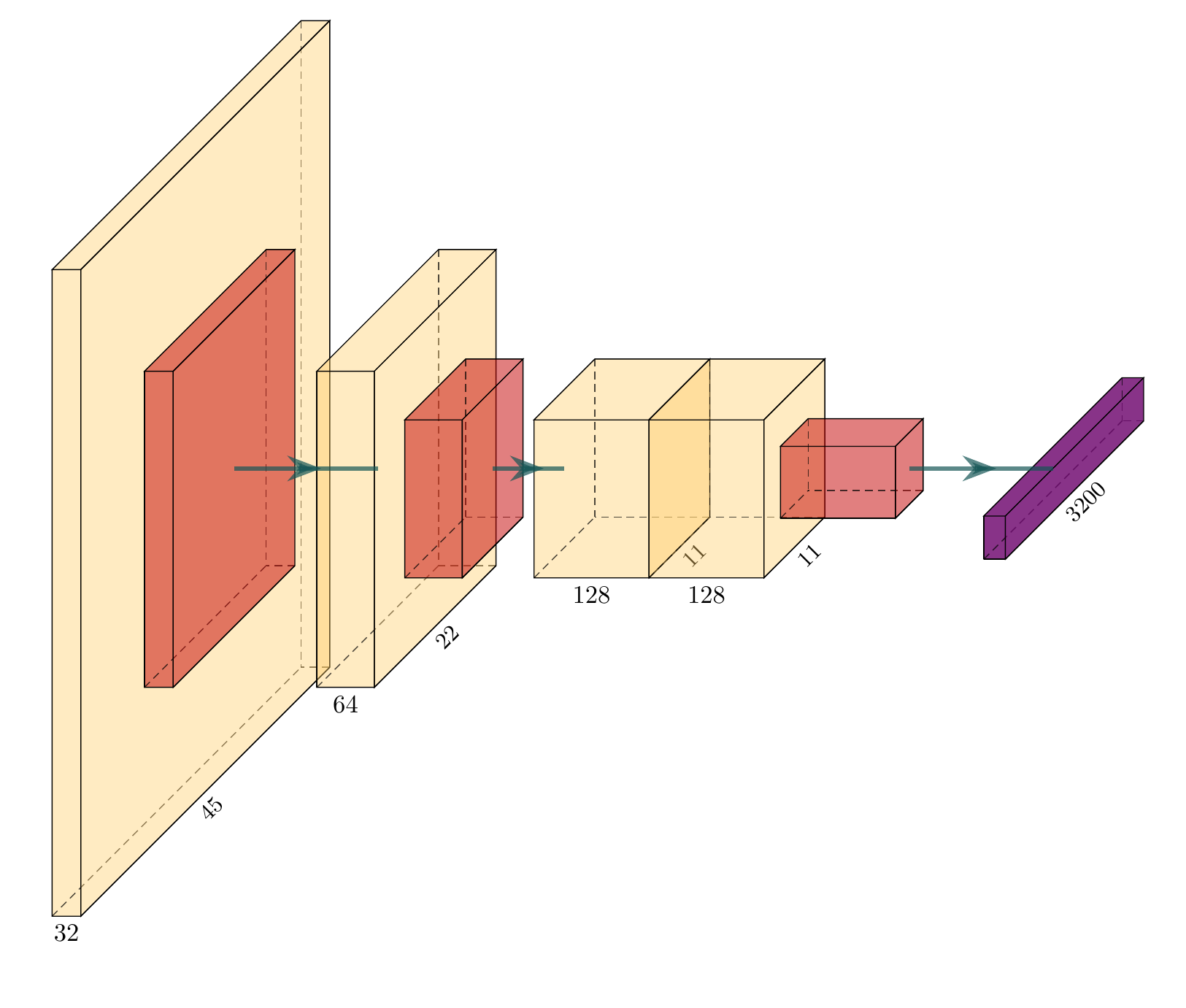} 
}\hfil
\subfloat[Fully connected layers\label{fig:nnim2}]{%
\includegraphics[scale=0.30]{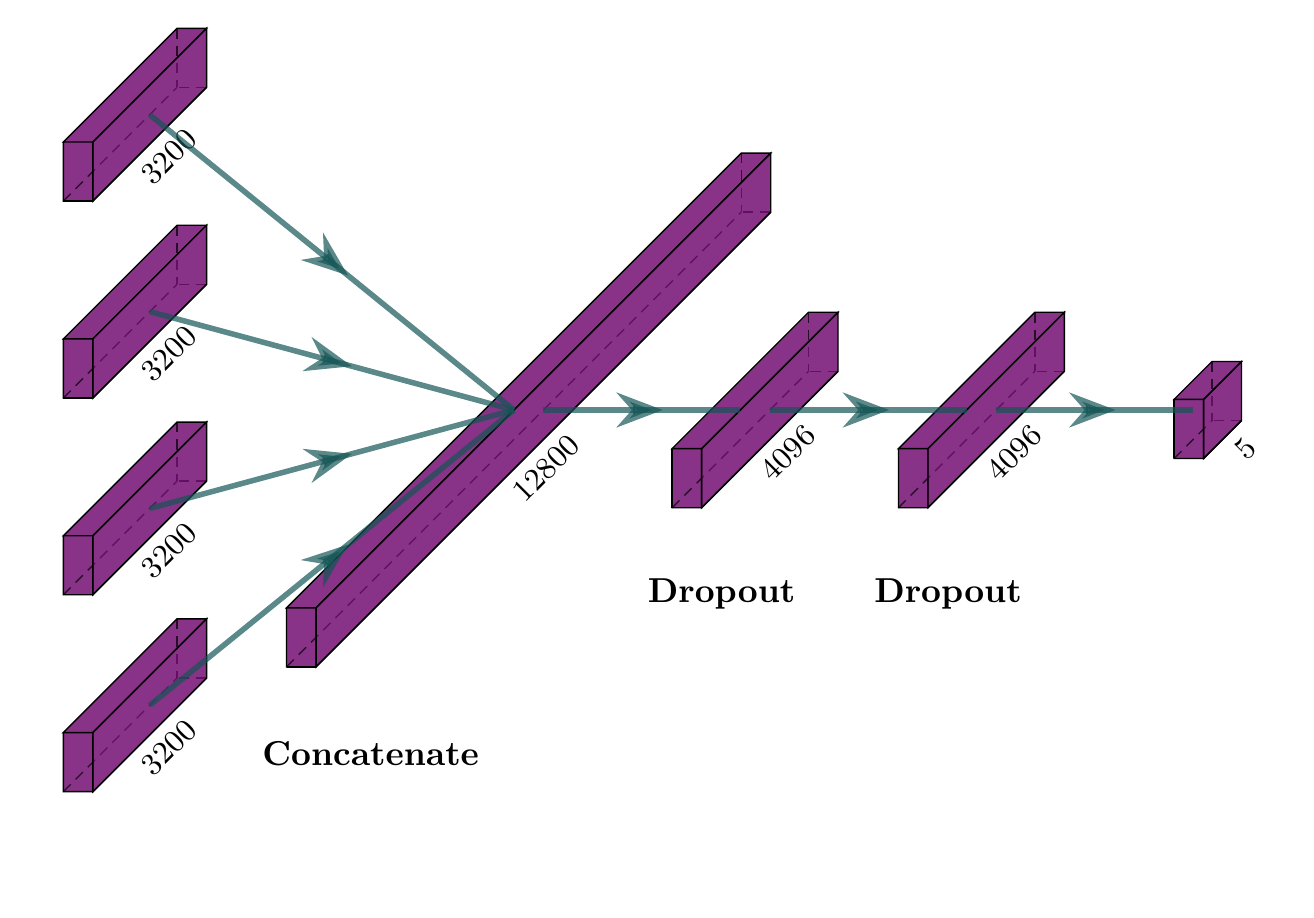}
}\hfil
\caption{Convolutional neural network architecture. \protect\subref{fig:nnim1} The input after augmentation has dimensions 45x45x1. Each convolutional block starts with a batch normalization layer and has a PReLU activation. The first convolutional layer is of dimension 45x45x32 with a 5x5 kernel size (in yellow), followed by a 2x2 Max-Pooling operation (in orange). The second convolutional layer is 22x22x64 with kernel size 3x3, followed by a 2x2 Max-Pooling operation. Then, we add two 11x11x128 convolutional layers with a 3x3 kernel that ends with a final 2x2 Max-Pooling operation, and the resulting feature maps are flattened into a 3200 fully connected layer (in purple). \protect\subref{fig:nnim2} Each augmented image gives a 3200 fully connected layer convolutional output (all augmented images share the same convolutional layers and filters), which are then concatenated into a 12800 fully connected layer. The two final layers have 4096 neurons in the case of an MVN regression, 2048 else; with Maxout activation \cite{goodfellow2013-maxout} and dropout with a rate of 0.5. The output layer has 5 neurons in the case of an MVN regression.}
\label{fig:architecture}
\end{figure}

The architecture of our network is inspired by the work of Dielman, who proposed a simple model specifically tuned for the Galaxy Zoo challenge, therefore adapted to our data \cite{ref_dielman_galaxy_2014}. Each image is augmented in four different parts by cropping thumbnails from high resolution images, centered on spatial modes of light profile. Then each augmented image is fed to the CNN. The complete architecture is explained in Fig.~\ref{fig:architecture}.
More details on the training process are explained in Section \ref{training}. Results obtained with this model are given in Section \ref{sec:res_1}.

\subsection{Estimation of blend related uncertainty}\label{sec:bdl}
 
As seen in Section \ref{sec:prob}, estimating the uncertainty due to the noise in the data is only one part of the problem. An estimated 60\% of the images represent blended scenes, for which a direct estimation of ellipticity does not make sense in the context of this work.
The uncertainty related to the blended images cannot be estimated simply with the variance of the MVN distribution. Indeed, in the case of a blended scene image, the network is not uncertain because of the noise but rather because this kind of images is not part of the training sample. This can be characterized by the epistemic uncertainty.

This uncertainty can be estimated using a Bayesian Neural Network (BNN), which assumes a probability distribution on the weights $W$ of the network instead of a single point estimate \cite{ref_hinton_keeping_1993}. Given a prior $p(w)$ on $W$ and a set $\mathcal{D}=\{(X_i,Y_i)\}_i$ of observations, the resulting posterior distribution $p(w|\mathcal{D}) \propto p(\mathcal{D} | w) \, p(w)$ is analytically impossible to compute. A variational Bayes optimization method is necessary to derive an approximate posterior $q_\theta(w)$ parameterized by hyper-parameters $\theta$. In MC dropout \cite{ref_gal_dropout_2016,ref_gal_phd_2016}, the considered search space includes all approximate posteriors resulting from applying \emph{dropout} \cite{ref_srivastava_dropout_2014}, i.e multiplying every neuron output (of selected layers) by an independent Bernoulli variable. The dropout rate is set to the conventional value of 0.5, as this leads to an approximate posterior that can achieve well calibrated uncertainty estimates \cite{ref_gal_dropout_2016}. However, there are other ways to define the posteriors such as dropout rate tuning \cite{DBLP:conf/nips/GalHK17}, or ensemble methods by training many networks \cite{DBLP:conf/nips/Lakshminarayanan17}. During training, standard stochastic gradient descent techniques can be used, thanks to the reparameterization trick, to search for an approximate posterior maximizing locally the ELBO \cite{ref_gal_dropout_2016}. During testing, the \emph{posterior predictive distribution} $p(Y|X,\mathcal{D})$ for some input $X$ can be estimated using Monte Carlo sampling: 
   \begin{equation}
   p(Y|X,\mathcal{D}) \approx \int p(Y|X,w)q_\theta(w)dw \approx \frac{1}{K} \sum_{k=1}^K p(Y|X,w_k)\,,
       \label{eq:marginal}
   \end{equation}
where $(w_k)_{k=1}^K \sim q_\theta(w)$ refer to weights of
$K$ independent dropout samples.

In the case of a multivariate regression problem like ours, every distribution $p(Y|X,w_k)$ is a MVN so that the resulting posterior predictive distribution in Eq.~\ref{eq:marginal} is a Gaussian mixture of order $K$. The uncertainty underlying this mixture can be summarized by its covariance matrix
$\Sigma_{pred.}(X,\mathcal{D}) = \mathrm{Cov}(Y|X, \mathcal{D})$.
This matrix accounts for both aleatoric and epistemic uncertainties, whose respective contributions can actually be separated in a way that generalizes the variance decomposition described in Depeweg \cite{ref_depeweg_uncertainty_2018}:
\begin{equation}
    \Sigma_{pred.}(X,\mathcal{D}) = \Sigma_{aleat.} (X,\mathcal{D}) + \Sigma_{epist.} (X,\mathcal{D})\, ,
\label{eq:predunc}
\end{equation}
where the first term represents the aleatoric uncertainty and can be computed as the mean of the covariance matrices for each of the $K$ output samples:
\begin{align}
   \Sigma_{aleat.} (X,\mathcal{D}) & = \mathbb{E}_{W|\mathcal{D}}(\mathrm{Cov}(Y|X,W))\, , \nonumber \\
   & \approx \frac{1}{K} \sum_{k=1}^K \Sigma(X,w_k) \, .
 \label{eq:datunc}
\end{align}
while the second represents the epistemic uncertainty and is estimated as the empirical covariance matrix of the $K$ mean vectors produced as outputs:
\begin{align}
   \Sigma_{epist.}(X,\mathcal{D}) & = \mathrm{Cov}_{W|\mathcal{D}}(\mathbb{E}(Y|X,W))\, , \nonumber \\
  & \approx \frac{1}{K} \sum_{k=1}^K (\mu(X,w_k)-\mu(X))(\mu(X,w_k)-\mu(X))^T \, .
\label{eq:modunc}
 \end{align}
where $\mu(X) = \frac{1}{K} \sum_{k=1}^K \mu(X,w_k)$.

Matrix $\Sigma_{epist.}(X,\mathcal{D})$ defines the epistemic uncertainty as the covariance matrix of the mean vectors over the posterior. This uncertainty will be high if the sampled predictions from each model vary considerably with respect to $W$. This would mean that no consistent answer can be deduced from the model and therefore it would be highly uncertain. 

Finally when the context requires to reduce these uncertainty matrices to uncertainty levels so that they can be compared, their determinants are used to define two corresponding scalar quantities:
 $$\mathcal{U}_{aleat.}(X,\mathcal{D}) = |\Sigma_{aleat.}(X,\mathcal{D})|\quad \text{and}\quad \mathcal{U}_{epist.}(X,\mathcal{D}) = |\Sigma_{epist.}(X,\mathcal{D})|\,.$$
    
\subsection{Training protocol}\label{training}

In order to train a BNN with an MVN output, the model needs to learn both the mean and the covariance matrix. The network's training diverges when trying to learn both at the same time, forcing us to separate the training into two steps. First, we train a simple neural network without an MVN output - we use only two output neurons representing the mean - using a L2 loss. Then, we transfer the filters of the convolutional layers into the model with an MVN output, but reinitialize the fully connected layers. This allows the model to converge smoothly as the mean of the MVN distribution has already been learned, allowing the covariance matrix to be calibrated accordingly. 


This protocol works well when training on noiseless images of isolated galaxies but fails when training on noisy images. Indeed, overfitting occurs during the training of the network without MVN. When transferring the filters to the MVN model, the mean of the MVN is not well calibrated enough and the training of the BNN diverges. To fix this, we adjust the protocol for the model without MVN, adding noise incrementally during training: we first submit noiseless images, and modify 5\% of the sample, switching from noiseless to noisy images, every 50 epochs for 1000 epochs. This prevents overfitting and allows the MVN model to converge after the transfer.

\section{Experiments}\label{sec:res}

\subsection{Estimation of uncertainty related to noise}\label{sec:res_1}

\begin{figure}[htbp]
\centering
\subfloat[Predicted ellipticity without noise\label{fig:classic_noiseless_galaxy}]{%
\quad\quad\quad\includegraphics[scale=0.30, trim=80 13 80 40, clip]{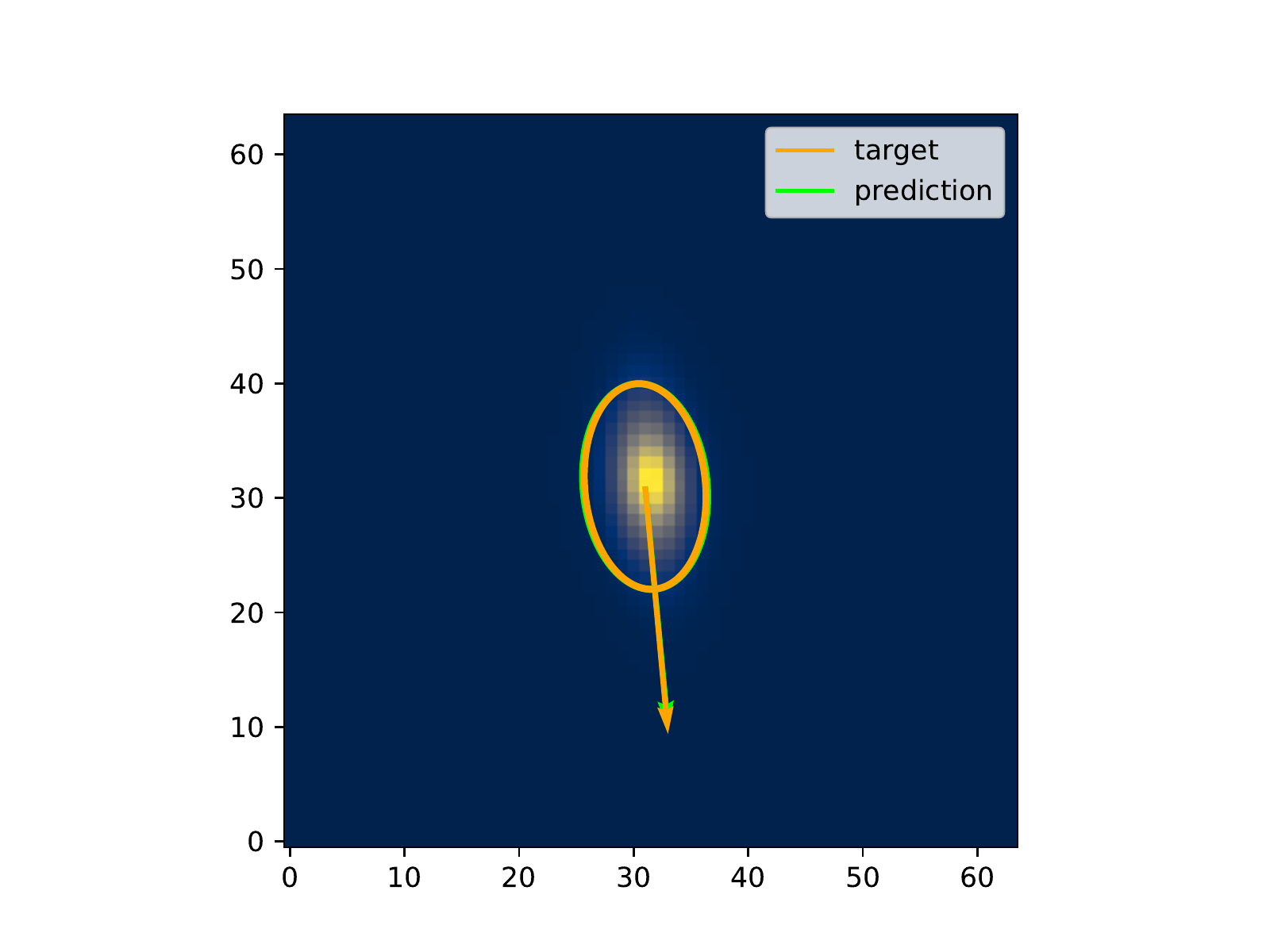} \quad\quad\quad
}\hfil
\subfloat[Predicted ellipticity with noise\label{fig:classic_noise_galaxy}]{%
\quad\quad\quad\includegraphics[scale=0.30, trim=80 13 80 40, clip]{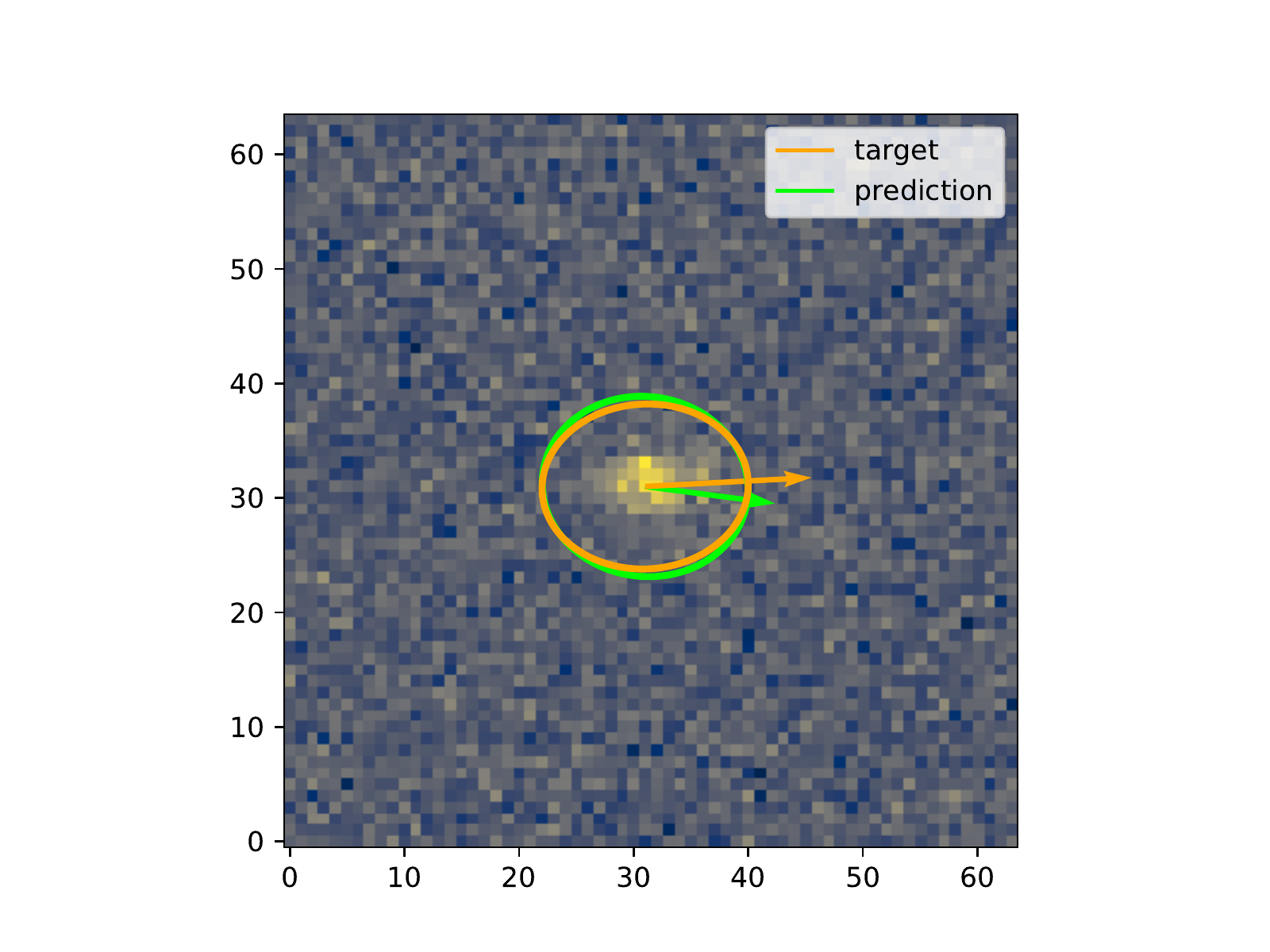}\quad\quad\quad
}\hfil
\caption{Galaxy images with the predicted ellipticity superimposed on them. The arrow and the corresponding elliptic shape are rendered in an arbitrary scale for visualization purposes. In orange: the true ellipticity. In green: the predicted ellipticity. \protect\subref{fig:classic_noiseless_galaxy} Galaxy without noise. \protect\subref{fig:classic_noise_galaxy} Galaxy with noise.}
\label{fig:classic_galaxy}
\end{figure}

\begin{figure}[htp]
\centering
\subfloat[Predicted ellipticities without noise\label{fig:classic_noiseless}]{%
\quad\quad\quad
\includegraphics[scale=0.33, trim=75 13 80 40, clip]{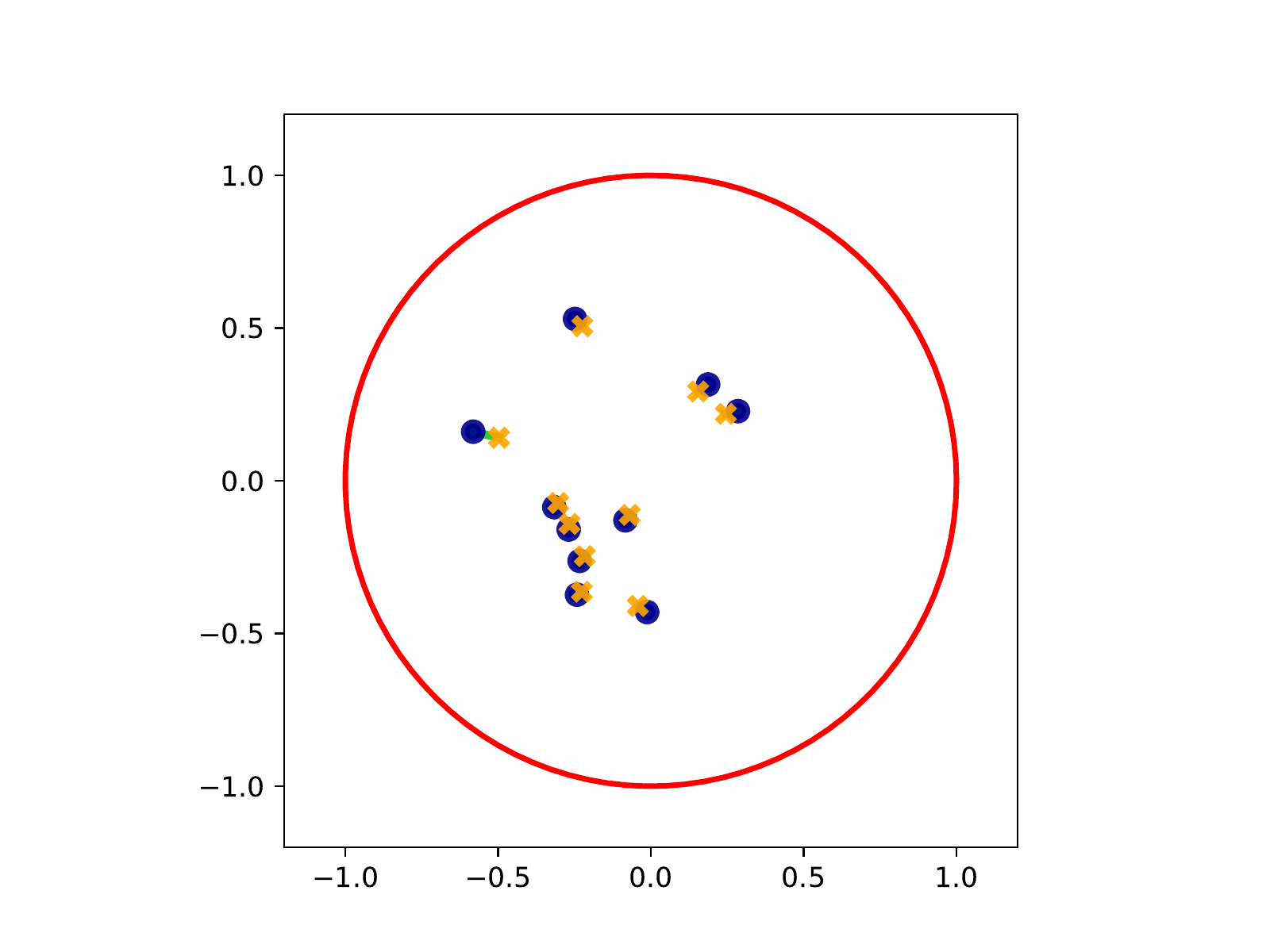} 
\quad\quad\quad
}\hfil
\subfloat[Predicted ellipticities with noise\label{fig:classic_noise}]{%
\quad\quad\quad
\includegraphics[scale=0.33, trim=75 13 80 40, clip]{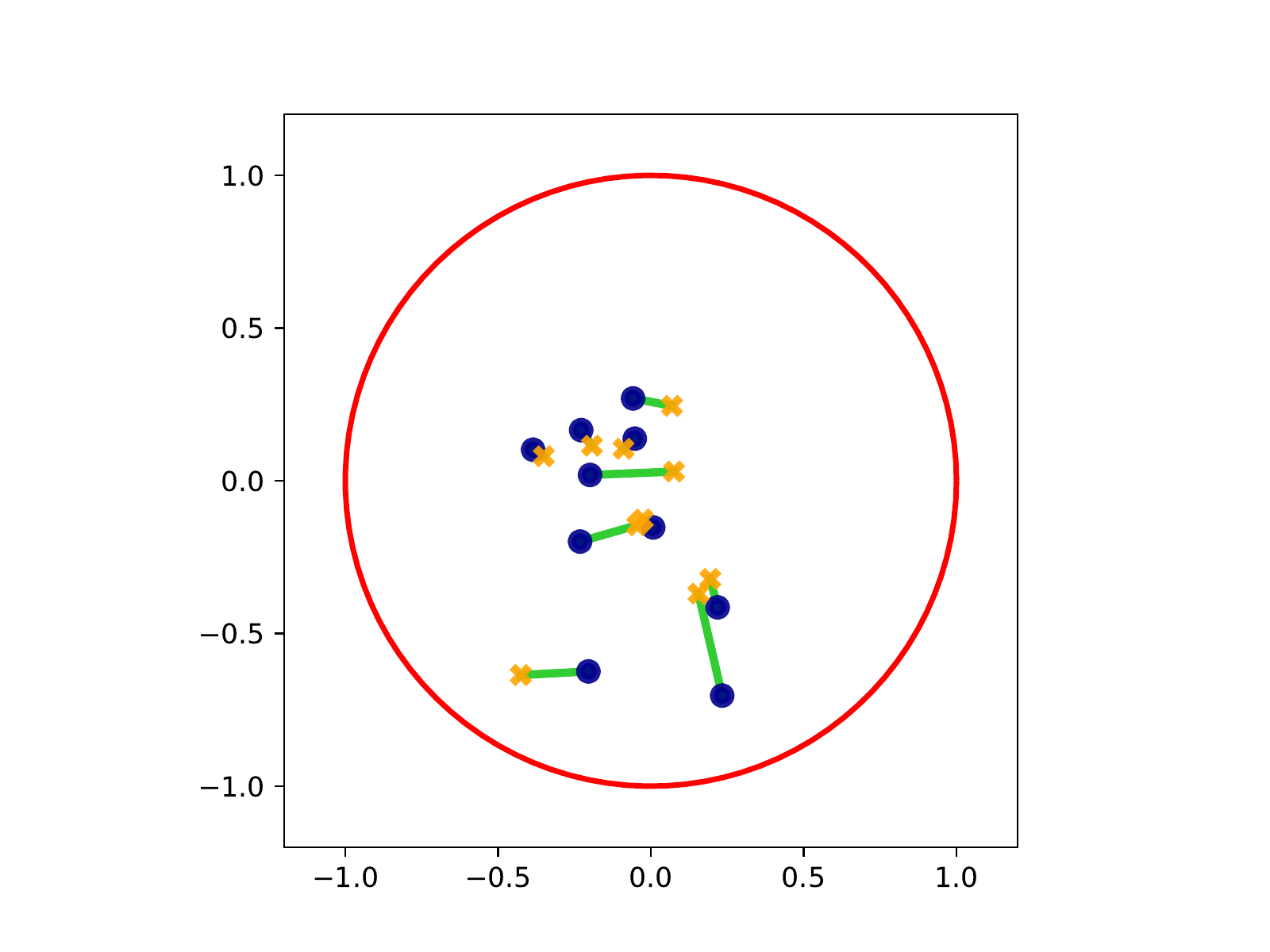}
\quad\quad\quad
}\hfil
\caption{Predicted ellipticities on the complex plane. In red: unit circle. In yellow: predicted ellipticities. In blue: target ellipticites. In green: difference between true and predicted values.}
\label{fig:classic}
\end{figure}

In this section we show that using an MVN as an output allows for a reliable and well calibrated estimation of the aleatoric uncertainty, i.e. uncertainty related to the noise in the data. 
In order to show that estimating the ellipticity of galaxies in the presence of background noise is complex and can induce incorrect predicted ellipticity values, we first train two simple CNNs without an MVN output: one on noiseless images and one on noisy images, accordingly tested on noiseless and noisy images respectively.
Figure \ref{fig:classic_galaxy} shows the images of galaxy with their target complex ellipticity superimposed, as well as the predicted one. 
The ellipse represents the estimated shape - with a fixed scale adapted for visualization - and the arrow is the corresponding complex ellipticity - modified with half its argument in order to be aligned wih the main axis of the ellipse. On this example, we can qualitatively see that the galaxy ellipticity on the noisy image is harder to estimate as the noise deforms the shape of the galaxy. Figure \ref{fig:classic} generalizes this observation as it shows a sample of the predicted ellipticities on the complex plane within the unit circle, with the target ellipticity and the difference between predicted and targeted values. 
While the model trained on noiseless data performs really well (Fig. \ref{fig:classic_noiseless}), it cannot achieve the same level of performance when trained on noisy data, losing part of its reliability (Fig. \ref{fig:classic_noise}). As such, using a simple CNN without any estimation of aleatoric uncertainty is not satisfying for our application.

\begin{figure}[htp]
\centering
\subfloat[Predicted ellipticities without noise\label{fig:Gaussian_noiseless}]{%
\quad\quad\quad
\includegraphics[scale=0.33, trim=75 13 80 40, clip]{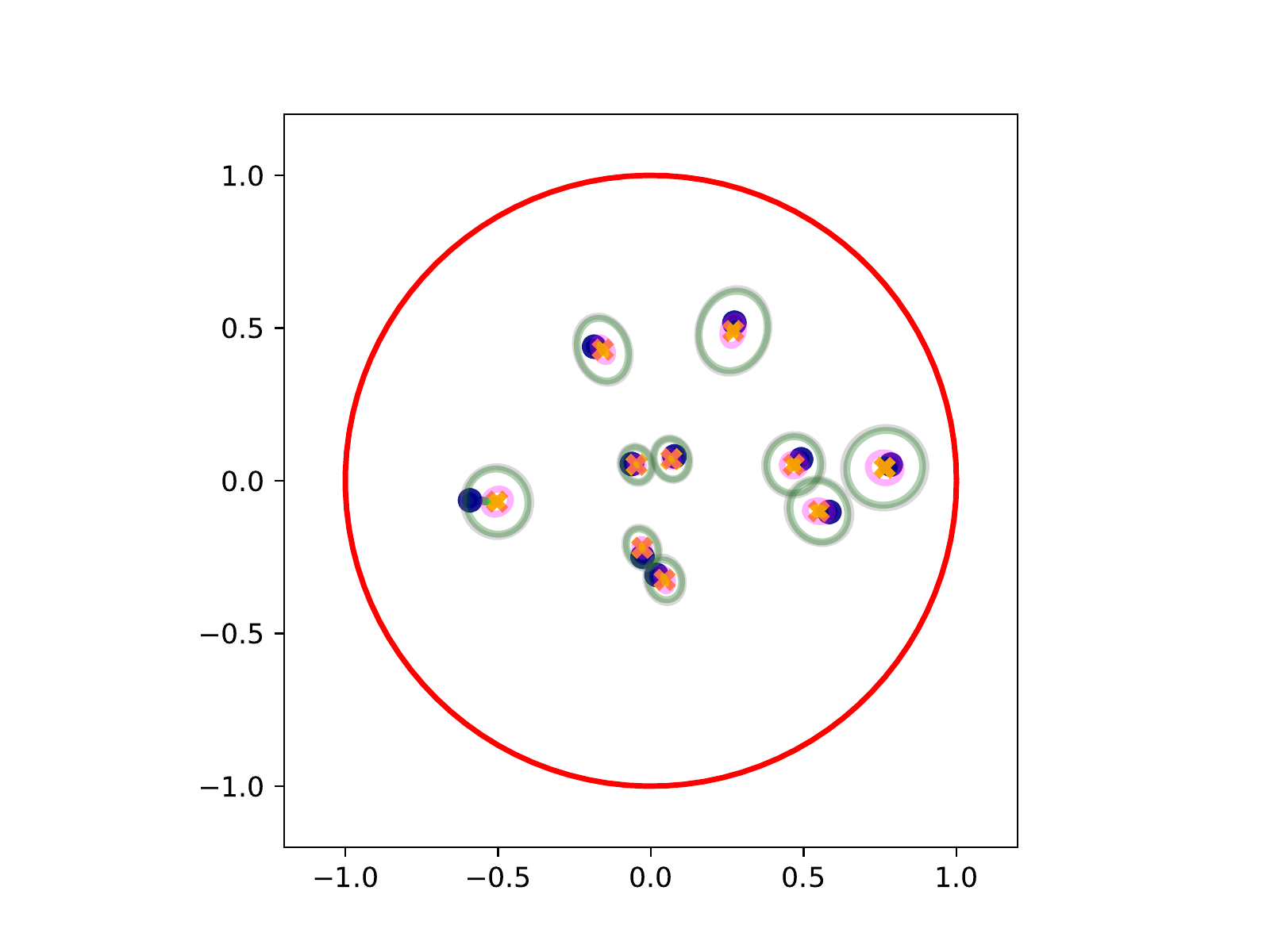} 
\quad\quad\quad
}\hfil
\subfloat[Predicted ellipticities with noise\label{fig:Gaussian_noise}]{%
\quad\quad\quad
\includegraphics[scale=0.33, trim=75 13 80 40, clip]{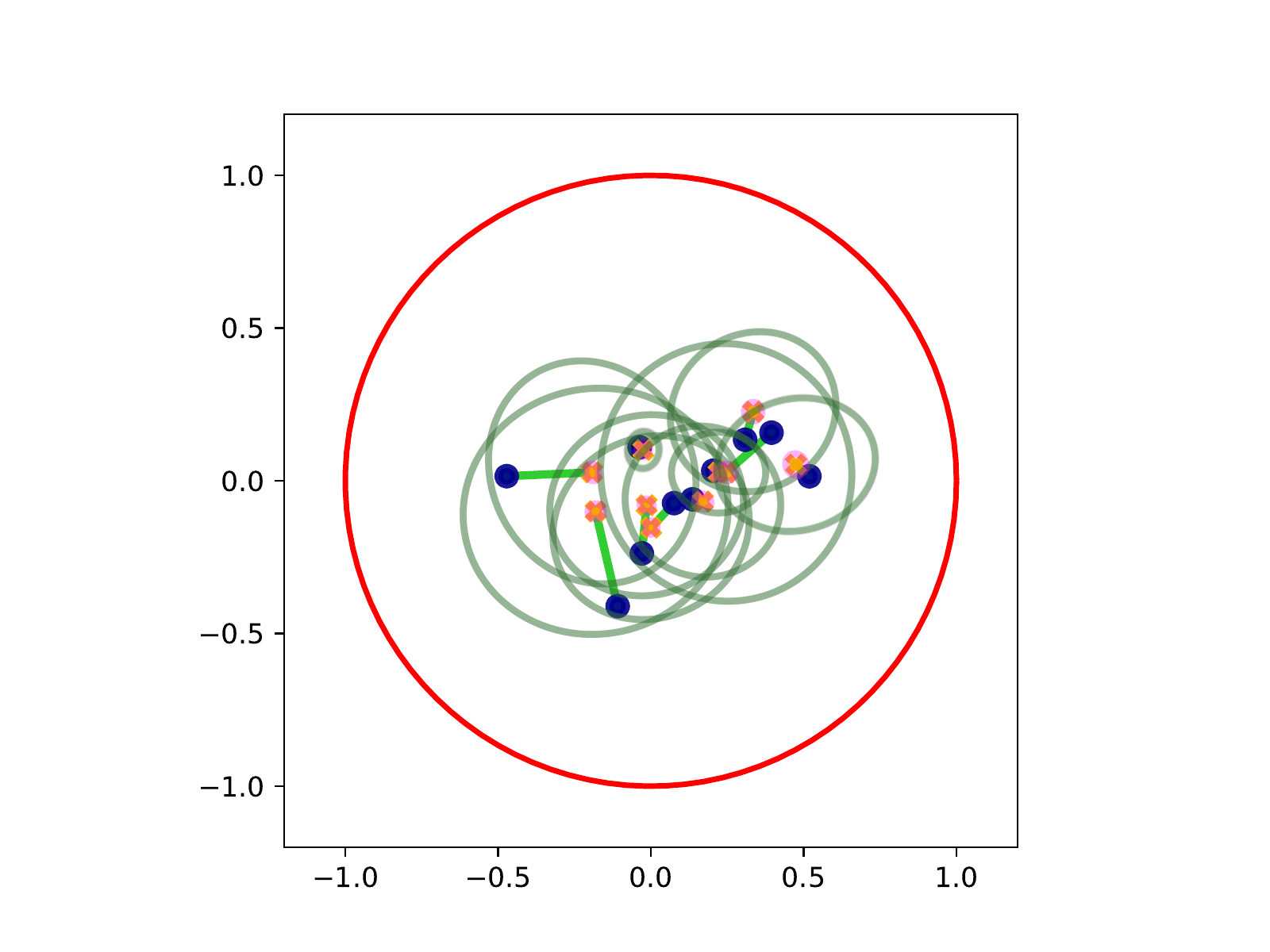}
\quad\quad\quad
}\hfil
\caption{Predicted ellipticities on the complex plane. In red: unit circle. In yellow: predicted ellipticities. In blue: target ellipticites. In light green: difference between true and predicted values. In pink: 90\% epistemic confidence ellipse. In dark green: 90\% aleatoric confidence ellipse. In grey: 90\% predictive confidence ellipse.}
\label{fig:Gaussian}
\end{figure}

 We now train two Bayesian Convolutional Neural Networks with an MVN distribution to estimate both epistemic and aleatoric uncertainties, as seen in Section \ref{sec:bdl}. Like the simple CNN models, we show in Fig. \ref{fig:Gaussian}, the ellipticities estimated from the BNNs on the complex plane. We also add the 90\% confidence ellipses of both epistemic, aleatoric and predictive uncertainties.
We observe that in both cases, the epistemic uncertainty is low if not negligible, meaning that the model is confident in its predictions. Put another way, all $K$ pairs of outputs $\mu(X,w_k)$ and $\Sigma(X,w_k)$ are roughly equal to their mean, respectively $\mu(X,w)$ and $\Sigma_{aleat.}(X,w)$, so that, according to Eq.~\ref{eq:predunc} and Eq.~\ref{eq:modunc} , $\Sigma_{epist.}(X,w) \approx 0$ and $\Sigma_{pred.}(X,w) \approx \Sigma_{aleat.}(X,w)$. The aleatoric uncertainty is low for noiseless images but higher for noisy ones, confirming that the noise corrupting galaxy images makes it more difficult for the model to consistently give an accurate ellipticity estimation.

\begin{figure}[htp]
\centering
\subfloat[Distribution of $z_1$\label{fig:z1_dist}]{%
\includegraphics[scale=0.33, trim=35 13 40 40, clip]{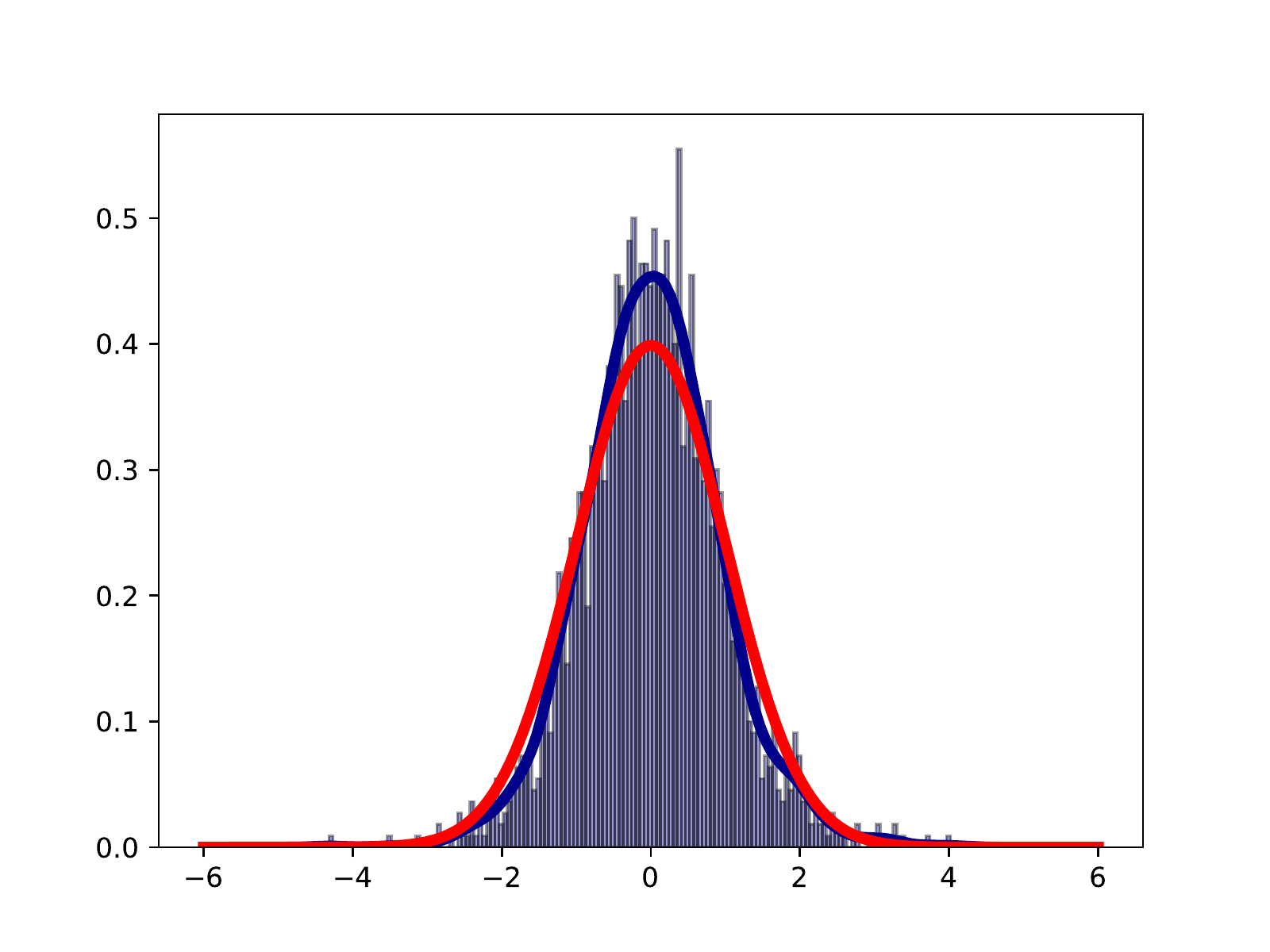} 
}\hfil
\subfloat[Distribution of $z_2$\label{fig:z2_dist}]{%
\includegraphics[scale=0.33, trim=35 13 40 40, clip]{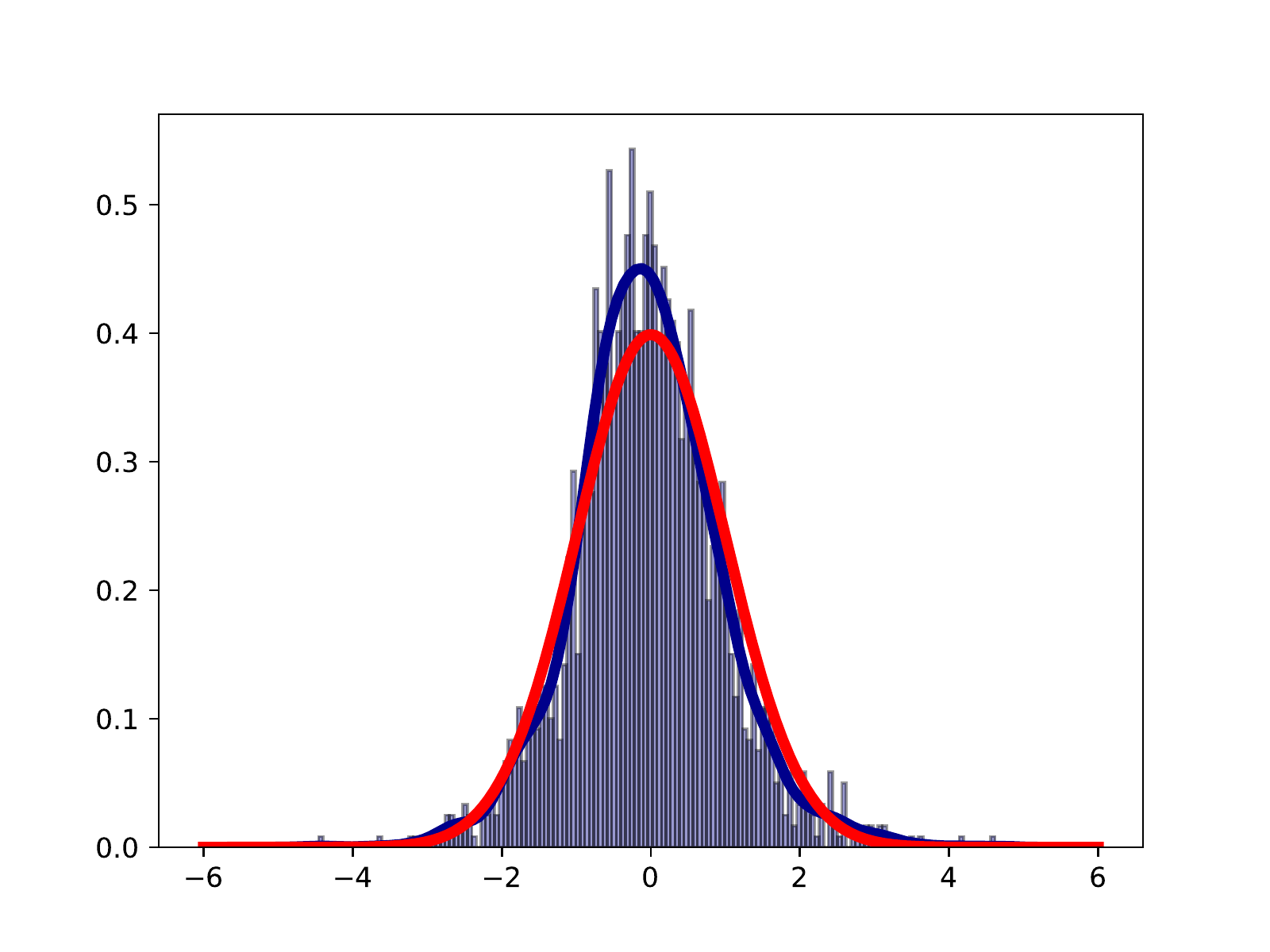}
}\hfil
\caption{Histogram of the standardized distributions on the  model trained with noisy images. In red: standard bell curve. In blue: histogram of the standardized distribution with the smoothed curve.}
\label{fig:standard_dist}
\end{figure}

Finally, in order to see if the MVN distribution is well calibrated, we standardize the output and check if the resulting distribution follows the standard distribution. More precisely, if we define:
\begin{equation}
    Z(X,w) = \Sigma_{pred.}(X,w)^{-\frac{1}{2}}(Y - \mu(X,w))\quad,
    \label{eq:standard_var}
\end{equation}
then the distributions of its two independent components $z_1 \sim Z(X,w)_1 $ and $z_2 \sim Z(X,w)_2 $ should be equivalent to the standard distribution $\mathcal{N}(0,1)$. Note that this is true only because all $K$ output MVNs are confounded. Figure \ref{fig:standard_dist} shows that the standardized distributions for the model trained on noisy images are indeed well calibrated and therefore the model is neither overestimating nor underestimating the predictive uncertainty.

\subsection{Estimation of uncertainty related to blending}\label{sec:res_2}

In the previous part we showed that our BNNs are well calibrated. Here we submit outliers to the networks in order to study the impact on epistemic uncertainty and whether it can be used to detect them.
Our models have only been trained on images of isolated galaxies, but astrophysical images can contain multiple overlapped galaxies. In that case, asking the model to measure a single ellipticity does not make sense. If the epistemic uncertainty behaves as expected, then its measurement would allow us to detect when a predicted ellipticity is incorrect due to the presence of multiple galaxies in the image.
We fed images of blended scenes to the two models trained on isolated galaxies (with or without noise), adding noise to the blended scenes only for the model trained on noisy images. 

\begin{figure}[h]
\centering
\subfloat[Predicted ellipticities, without noise \label{fig:blended_noiseless}]{%
\quad\quad\quad\quad
\includegraphics[scale=0.33, trim=75 13 80 40, clip]{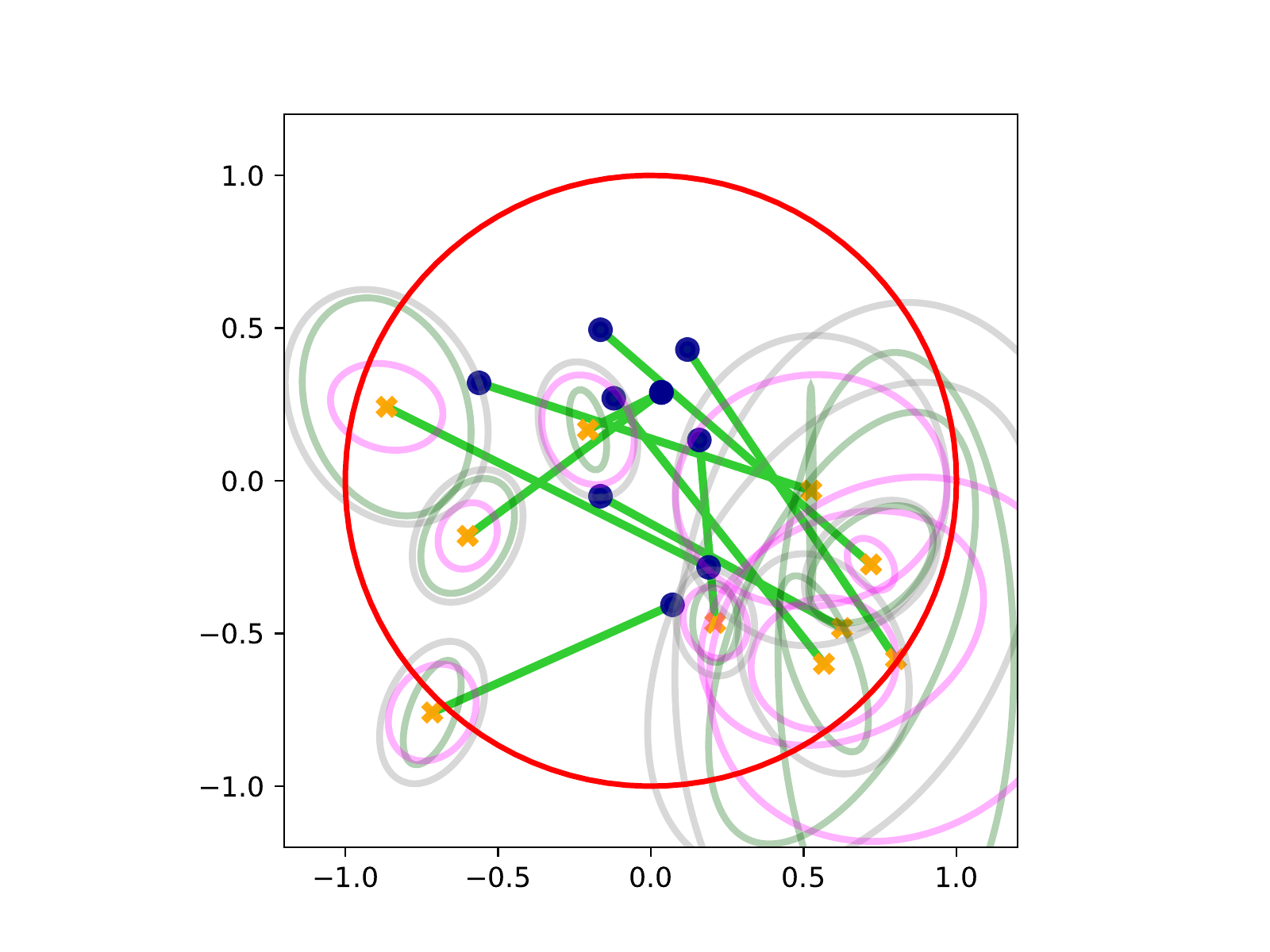} 
\quad\quad\quad\quad
}\hfil
\subfloat[Predicted ellipticities, with noise \label{fig:blended_noise}]{%
\quad\quad\quad
\includegraphics[scale=0.33, trim=75 13 80 40, clip]{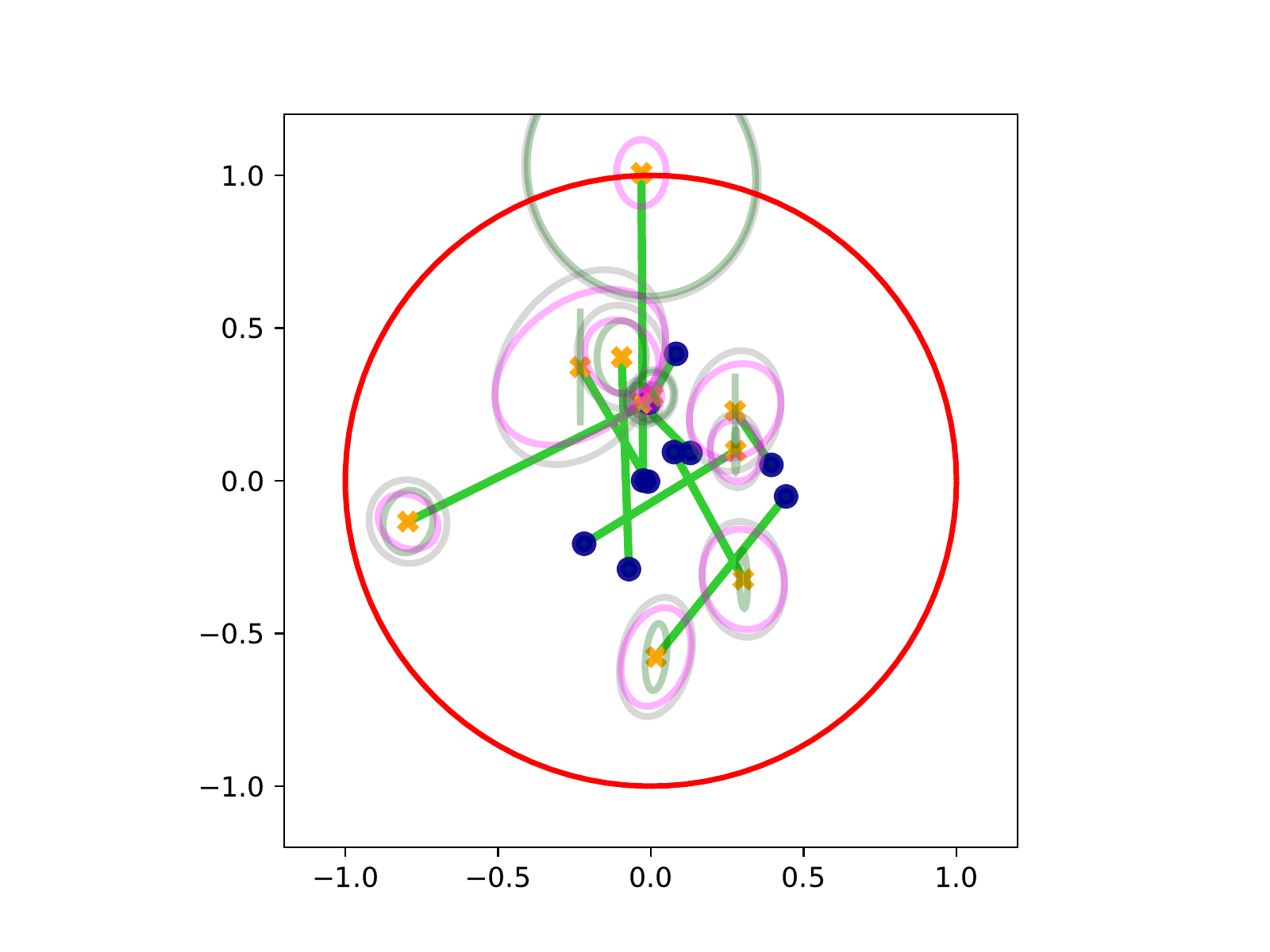}
\quad\quad\quad
}\hfil
\caption{Predicted ellipticities on the complex plane for blended galaxies images. In red: unit circle. In yellow: predicted ellipticities. In blue: target ellipticites (label of the centered galaxy). In light green: difference between true and predicted values. In pink: 90\% epistemic confidence ellipse. In dark green: 90\% aleatoric confidence ellipse. In grey: 90\% predictive confidence ellipse.}
\label{fig:blended_ellipticities}
\end{figure}

\begin{figure}[h]
\centering
\subfloat[Blended galaxies without noise \label{fig:blended_galaxy_noiseless}]{%
\quad\quad
\includegraphics[scale=0.33, trim=80 13 80 13, clip]{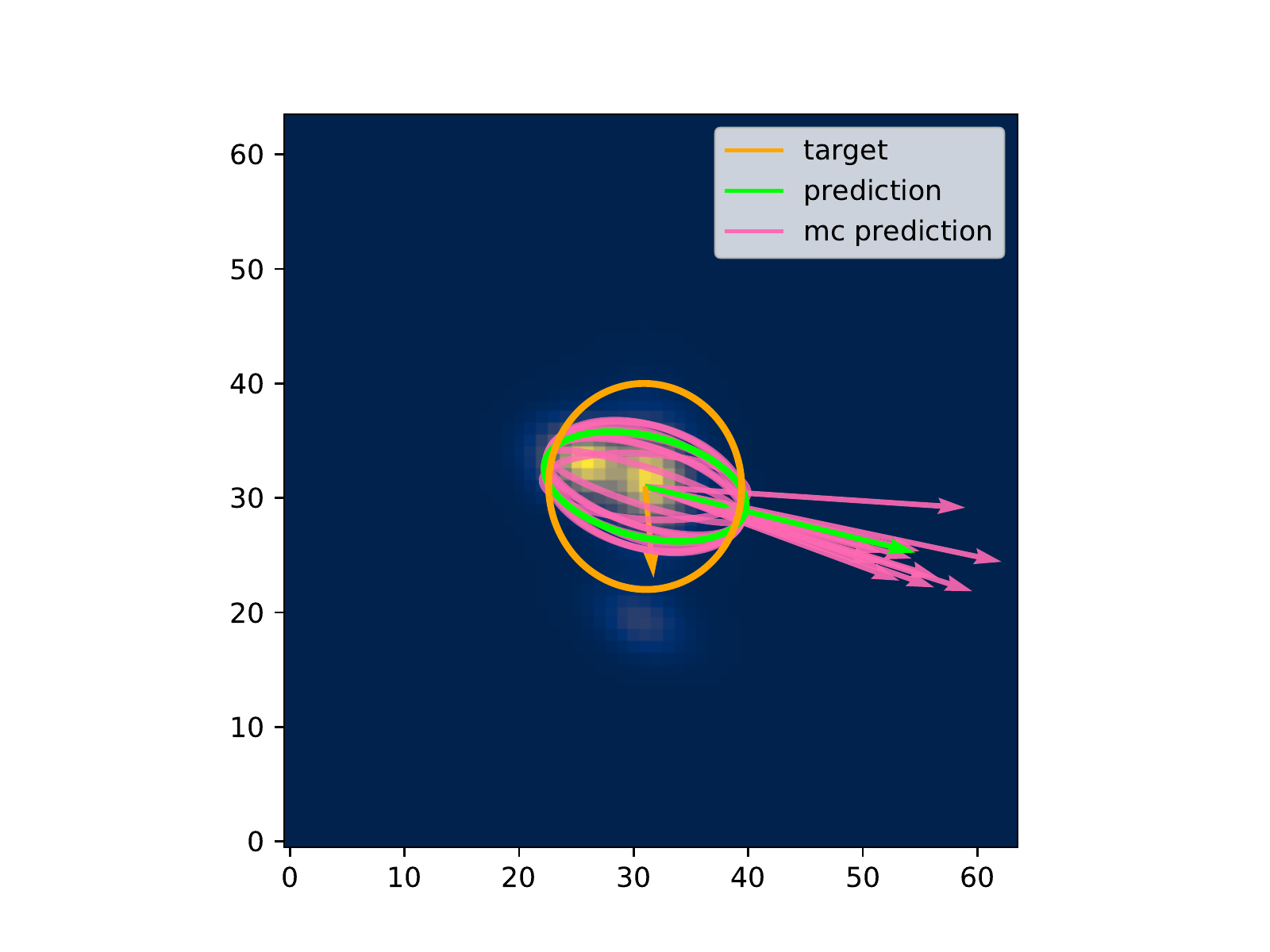} 
\quad\quad
}\hfil
\subfloat[Blended galaxies with noise\label{fig:blended_galaxy_noise}]{%
\quad\quad
\includegraphics[scale=0.33, trim=80 13 80 13, clip]{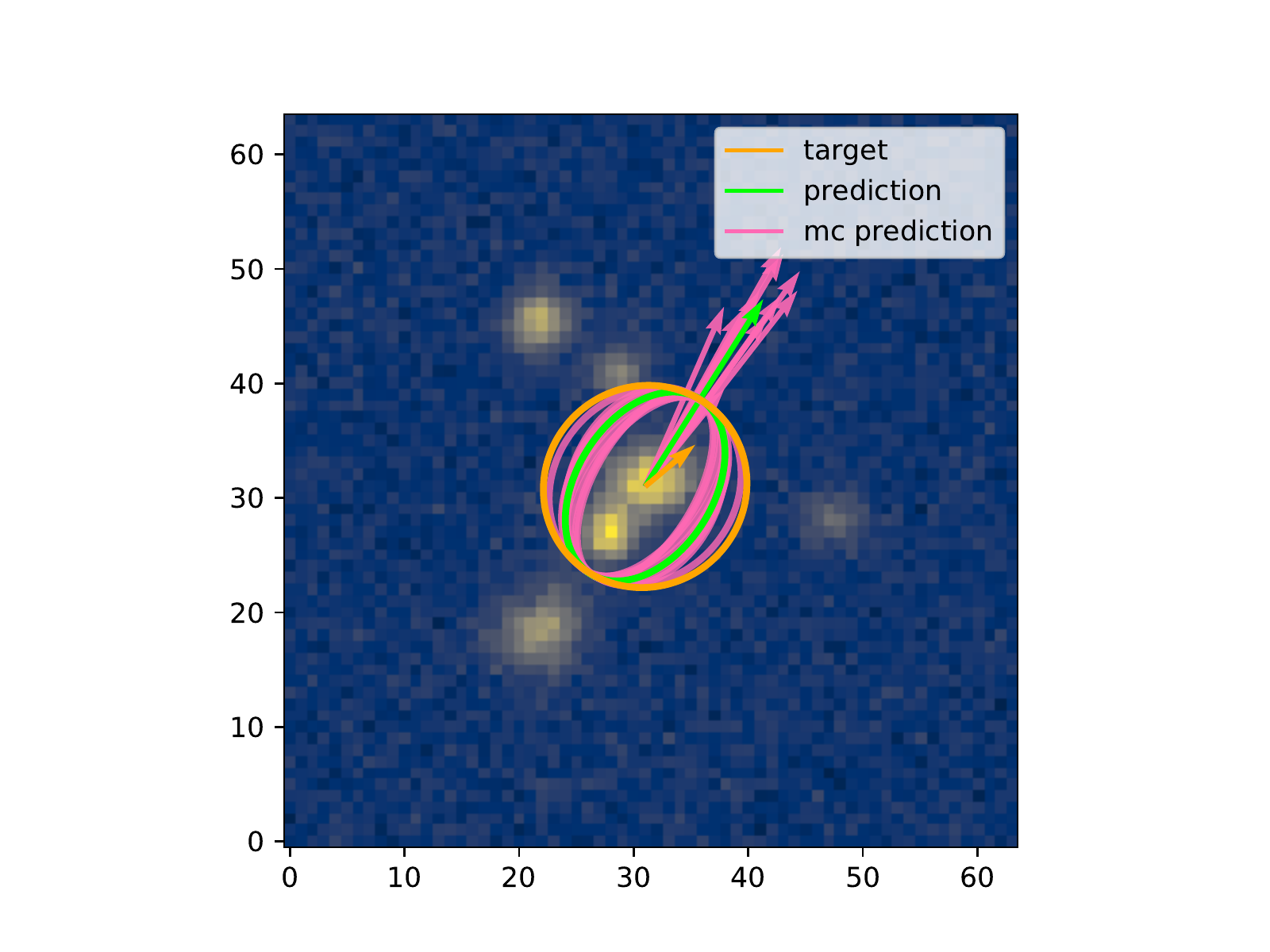}
\quad\quad
}\hfil
\caption{Blended galaxies images with the predicted ellipticity superimposed on them. The arrow and the corresponding elliptic shape are rendered in an arbitrary scale for visualization purposes. In orange: the true ellipticity (label for the galaxy in the center). In green: the predicted ellipticity. In pink: the individual MC dropout predicted ellipses. The green ellipticity is therefore the mean of the pink ones. On both images the prediction is uncertain as the individual MC samples slightly diverge from the mean.}
\label{fig:blended_galaxies}
\end{figure}

Results shown in Fig. \ref{fig:blended_ellipticities} demonstrate that in both cases the predictions are particularly inexact when compared to the target ellipticity of the central galaxy. Also, and as expected, the epistemic uncertainty is much higher for these blended scenes than for isolated galaxy images. However, the aleatoric uncertainty gives incoherent values as the model has not been trained to evaluate it on blended images: notice how the aleatoric ellipses are more flattened with a lower area.
Figure \ref{fig:blended_galaxies} permits to visualise the behavior of the epistemic uncertainty. It shows how the ellipticities sampled with dropout slightly diverge compared to the mean prediction. Here the model cannot give a consistent answer and therefore its prediction should be deemed untrustworthy.

\begin{figure}[h]
\centering
\subfloat[ROC curve, model without noise \label{fig:roc_curve_noiseless}]{%
\quad\quad
\includegraphics[scale=0.33, trim=50 8 80 40, clip]{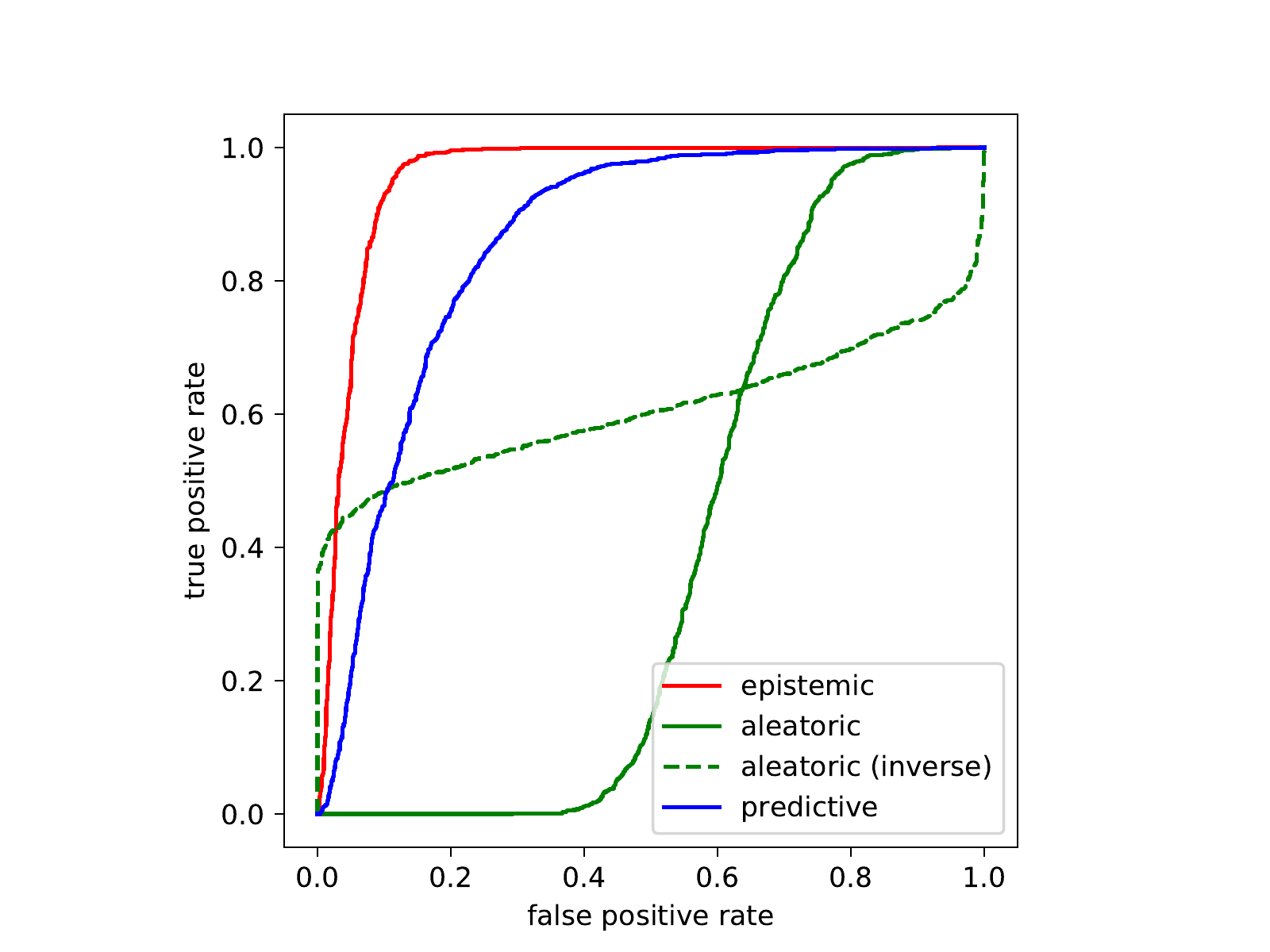} 
\quad\quad
}\hfil
\subfloat[ROC curve, model with noise\label{fig:roc_curve_noise}]{%
\quad
\includegraphics[scale=0.33, trim=50 8 80 40, clip]{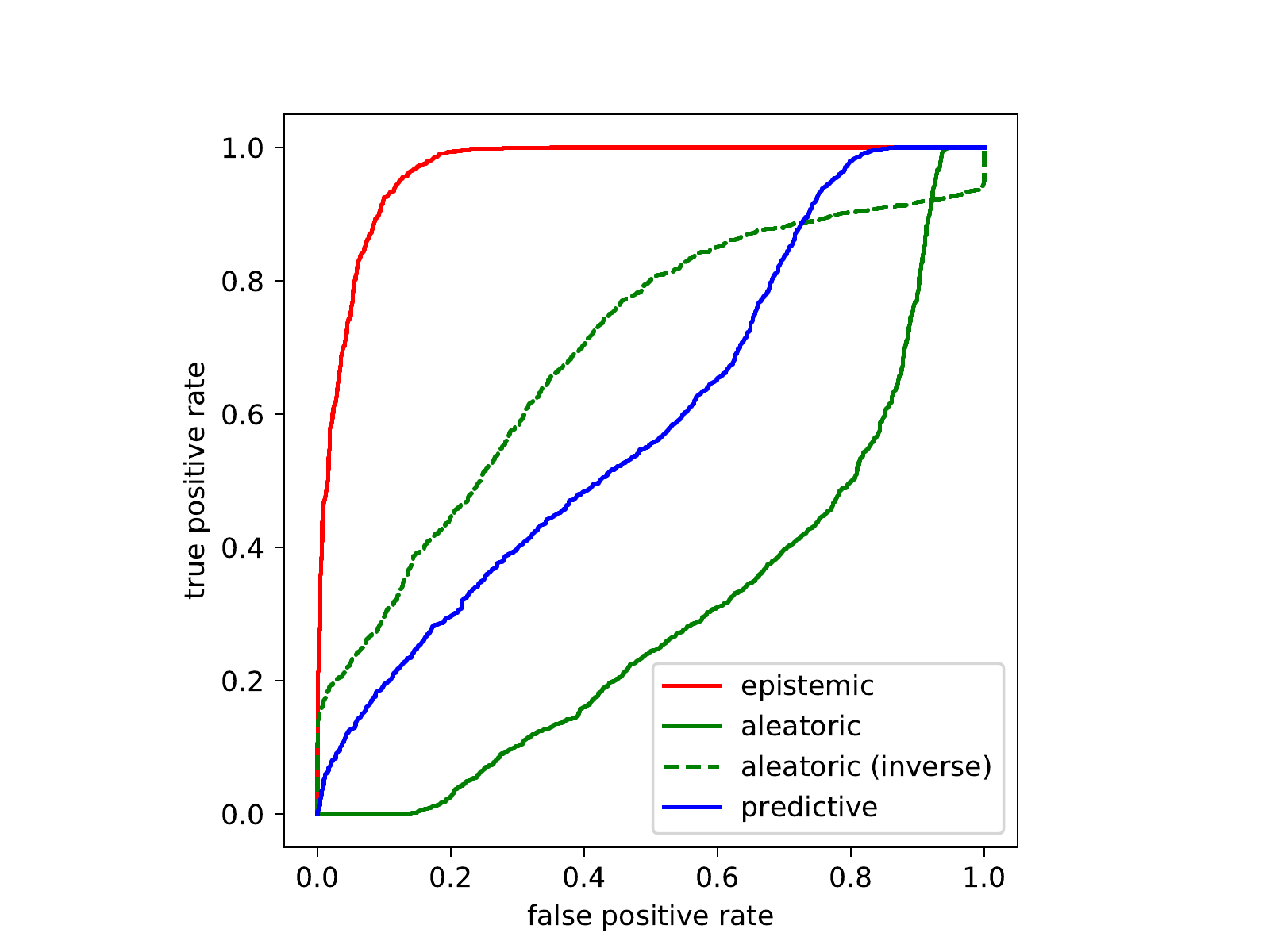}
\quad
}\hfil
\caption{ROC curves for detecting outliers for aleatoric, epistemic and predictive uncertainty. \protect\subref{fig:roc_curve_noiseless} ROC curve, model without noise. \protect\subref{fig:roc_curve_noise} ROC curve model with noise. Since the aleatoric ROC curve gives incoherent answers on outliers (see Fig. \ref{fig:blended_ellipticities}), we also plot the complementary classifier as a dashed line.}
\label{fig:roc_curve}
\end{figure}

To quantify the quality of the epistemic uncertainty when it comes to detecting incoherent predictions due to outliers, we computed the ROC curves for each uncertainty type. More precisely, we reduce each covariance matrix (aleatoric, epistemic and predictive) to a scalar by computing its determinant. We interpret these estimates as a scoring function to assess whether an image is an outlier, i.e. a blended image: the higher the score, the more likely the image contains a blend. Finally, we compute for each of these scoring functions its ROC curve. We repeat that process for both networks trained with noisy and noiseless data. The results are shown in Fig. \ref{fig:roc_curve}.

These ROC curves are also summarized by their associated Area Under Curve (AUC) on Tab.~\ref{tab:AUC_values}. The epistemic uncertainty clearly appears as the most consistent ``metric'' to detect outliers and therefore to give useful information about the confidence in the model predictions. Even the predictive uncertainty performs worse than the epistemic one. This is especially true in the presence of noise since the aleatoric uncertainty then occupies a more important part of the predictive one compared to the noiseless case. Notice that the aleatoric ROC curve is mostly below the diagonal with an AUC below 0.5, meaning it performs worse than a random classifier. This is due to the fact that the model has not been trained to evaluate aleatoric uncertainty on blended scenes. As seen in Fig. \ref{fig:blended_ellipticities}, the aleatoric ellipses are more flattened in the blended cases, meaning its determinant is lower. Thus the aleatoric uncertainty is on average lower on blended scenes when compared to isolated ones. 
To compensate, results of the complementary classifier for the aleatoric uncertainty are shown. It is still not as satisfying as the epistemic uncertainty.
While using epistemic uncertainty to identify inconsistent predictions due to a lack of knowledge is highly effective, we note that few blended images still have low epistemic uncertainty due, for instance, to a large galaxy that obstructs all of the other ones, making the image actually closer to an isolated galaxy image.

\begin{figure}[h]
\centering
\subfloat[Mean error curve, model without noise \label{fig:error_curve_noiseless}]{%
\quad\quad
\includegraphics[scale=0.33, trim=15 8 40 35, clip]{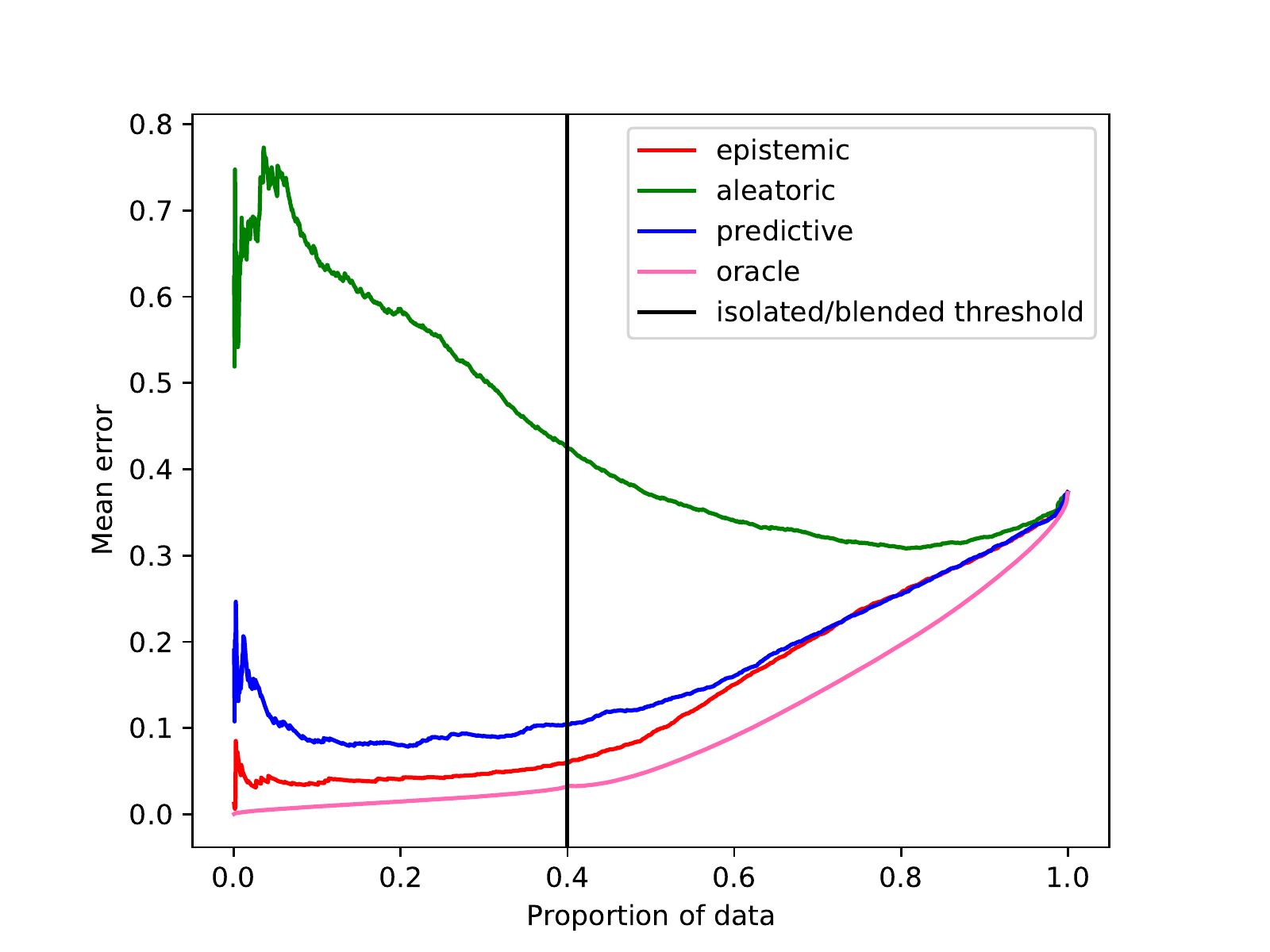} 
\quad\quad
}\hfil
\subfloat[Mean error curve, model with noise\label{fig:error_curve_noise}]{%
\quad
\includegraphics[scale=0.33, trim=15 8 40 35, clip]{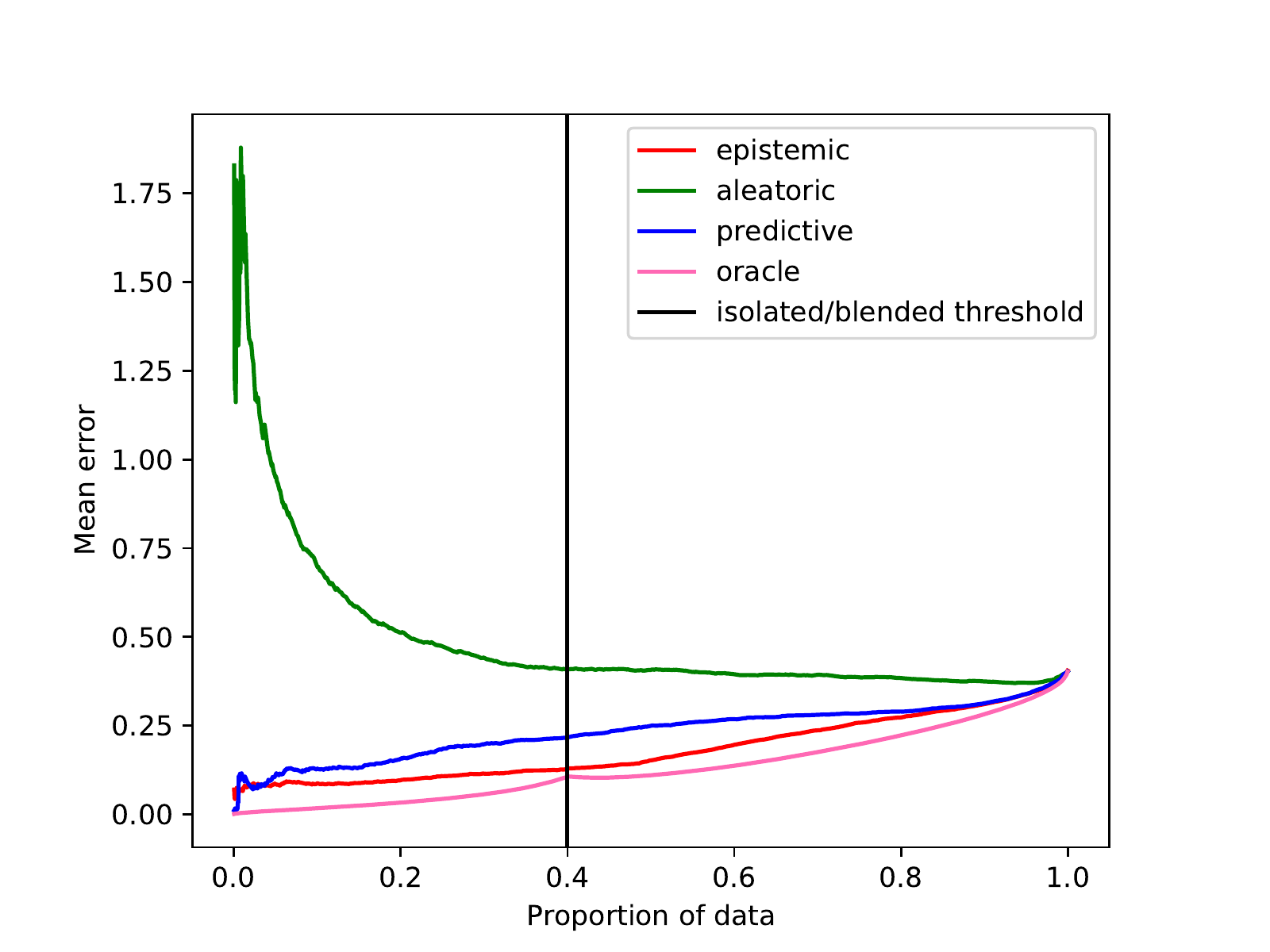}
\quad
}\hfil
\caption{Mean error curves w.r.t. data proportion for aleatoric, epistemic and predictive uncertainty. \protect\subref{fig:error_curve_noiseless} Mean error curve without noise. \protect\subref{fig:error_curve_noise} Mean error curve with noise. In black the threshold between the proportion of isolated galaxies: $[0,0.4]$ and blended galaxies: $[0.4,1]$. In pink the oracle curve, where the data is sorted by the predictive error. The closest a curve is to the oracle the better.}
\label{fig:error_curve}
\end{figure}

Finally, we evaluate how each type of uncertainty is a reliable representation of the risk of error in ellipticity prediction. Unfortunately, in the presence of blended images, the predictive distribution is no longer a simple MVN but a mixture of K well separated Gaussian distributions. The normalization process that allowed us to obtain the results presented in Fig. \ref{fig:standard_dist} is no longer applicable here. It is still possible to study the relationship between the uncertainty and the ellipticity prediction error testing a trivial rule: the higher the uncertainty, the more important we expect the error to be. To do so, we do three sorting of the images according to each uncertainty type, from the lowest uncertainty to the highest, on a scale from $0$ to $0.4$ for isolated objects, and from $0.4$ to $1$ for blended scenes. We then compute the mean ellipticity error considering the proportion of the sorted data from 0 to 1. For blended scenes the ellipticity prediction error is computed w.r.t. the ellipticity of the centered galaxy. We repeat this experiment twice, for networks trained on noiseless and noisy data. Finally we add an "oracle" curve where the data is sorted directly according to the ellipticity prediction error which represents a perfect sorting. Results are shown in Fig. \ref{fig:error_curve}.

\begin{table}[h]
    \centering
    \scalebox{1}{
    \begin{tabular}{|l|l|l|}
    \hline
    Uncertainty &  AUC noiseless & AUC noise \\
    \hline
    Epistemic  & {\bfseries 0.956} &  {\bfseries 0.969}\\
    Aleatoric & 0.394 & 0.306\\
    Aleatoric (inverse) & 0.606 & 0.694\\
    Predictive & 0.856 & 0.594\\
    \hline
    \end{tabular}
    }
    \vspace{0.3cm}
    \caption{AUC values for all uncertainties, for the model with  and without noise. Here, the epistemic uncertainty is clearly the best to detect outliers, as its AUC value is close to 1 in both noisy and noiseless datasets.}
    \label{tab:AUC_values}
\end{table}

Once again, epistemic uncertainty proves to be best suited to anticipate ellipticity predictive error. The samples with the lowest epistemic uncertainty have the lowest mean ellipticity error and conversely, while samples with low aleatoric uncertainty can already have high mean error. 
Consequently, on real astrophysical data, when the predictive ellipticity error is obviously unknown, relying on the epistemic uncertainty to reject, or minimize the impact of, a sample because of its probable predictive error is the best way to go.

\section{Conclusion}\label{sec:conclusion}


We developed a Bayesian approach to estimate the posterior distribution of galaxy shape parameters using convolutional neural networks and MC-Dropout. In addition to a precise measurement of the ellipticities, this approach provides a calibrated estimation of the aleatoric uncertainty as well as an estimation of the epistemic uncertainty. We showed that the latter is behaving according to expectations when applied to different kind of galaxy images, and is well-suited to identify outliers and to anticipate high predictive ellipticity error. These results confirm the suitability of Bayesian neural networks for galaxy shape estimation and incite us to continue exploring their use to go from ellipticity posterior distributions, estimated from multi-band galaxy images, to cosmic shear estimation.

\subsubsection{Acknowledgements}\hfill\\
The first author is preparing a PhD thesis at the LORIA Lab in the context of the AstroDeep Research Project (\url{https://astrodeep.pages.in2p3.fr/website/projects/}) funded by ANR under the grant ANR-19-CE23-0024.

%
%
%
%

\end{document}